\documentclass{article}
\usepackage{jfrExamplee}
\usepackage{graphicx}
\usepackage{apalike}
\usepackage{setspace}
\usepackage{graphics} %
\usepackage{epsfig} %
\usepackage{leftidx}
\usepackage{comment}
\usepackage{physics}
\usepackage{subcaption}
\usepackage{float}
\usepackage{amsfonts}
\usepackage{multirow}
\usepackage{color}
\usepackage{array}
\usepackage{tabularx}
\usepackage{tikz}
\usepackage{appendix}
\usetikzlibrary{shapes.geometric, arrows}
\usetikzlibrary{shadows}
\usepackage{hyperref}
\hypersetup{
    colorlinks=true,
    linkcolor=blue,
    filecolor=magenta,      
    urlcolor=cyan,
}
\usepackage[section]{placeins}
\usepackage[normalem]{ulem}

\newcommand{\eg}{\emph{e.g.},}
\newcommand{\ie}{\emph{i.e.},}

\newcommand{\cred}[1]{{\color{black}#1}}
\newcommand{\cblue}[1]{{\color{black}#1}}

\title{
Canopy Density Estimation in Perennial Horticulture Crops Using 3D Spinning Lidar SLAM}
\author{Thomas Lowe
\And
Peyman Moghadam\thanks{Corresponding author: Dr. Peyman Moghadam, \emph{Peyman.Moghadam}@csiro.au Tel: +61733274601 }
\And
Everard Edwards 
\And
Jason Williams
\thanks{Thomas Lowe, Peyman Moghadam and Jason Williams are with the Robotics and Autonomous Systems, CSIRO, DATA61, Brisbane, QLD 4069, Australia. Everard Edwards is with the CSIRO Agriculture \& Food, Waite Campus, SA 5064,  Australia.
E-mails: {\tt\small \emph{Thomas.Lowe}@csiro.au,  \emph{Peyman.Moghadam}@csiro.au,  \emph{Everard.Edwards}@csiro.au,  \emph{Jason.Williams}@csiro.au}}
}

\begin{document}

\maketitle

\begin{abstract}

\cblue{We propose a novel, canopy density estimation solution using a 3D ray cloud representation for perennial horticultural crops at the field scale. To attain high spatial and temporal fidelity in field conditions, we propose the application of continuous-time 3D SLAM (Simultaneous Localisation and Mapping) to a spinning lidar payload (AgScan3D) mounted on a moving farm vehicle.} The AgScan3D data is processed through a Continuous-Time SLAM algorithm into a globally registered 3D ray cloud. The global ray cloud is a canonical data format (a digital twin) from which we can compare vineyard snapshots over multiple times within a season and across seasons. Then, the vineyard rows are automatically extracted from the ray cloud and a novel density calculation is performed to estimate the maximum likelihood canopy densities of the vineyard. This combination of digital twinning, together with the accurate extraction of canopy structure information, allows entire vineyards to be analysed and compared, across the growing season and from year to year. The proposed method is evaluated both in simulation and field experiments. Field experiments were performed at four sites, which varied in vineyard structure and vine management, over two growing seasons \cblue{and 64 data collection campaigns}, resulting in a total traversal of 160 kilometres, 42.4 scanned hectares of vines with a combined total of approximately 93,000 scanned vines. \cblue{Our experiments show canopy density repeatability of 3.8\% (Relative RMSE) per vineyard panel, for acquisition speeds of 5-6 km/h, and under half the standard deviation in estimated densities when compared to an industry standard gap-fraction based solution.} The code and field datasets are available at
\cblue{\href{https://github.com/csiro-robotics/agscan3d}{https://github.com/csiro-robotics/agscan3d}.}

\end{abstract}
\section{Introduction} 
\label{sec:intro}

Canopy management plays a key role in most, if not all, perennial horticulture crops, primarily to optimise the balance between the fruit yield and photosynthetic productivity (\eg{} \cite{KliewerWeaver,Reynolds}) and to provide the appropriate micro-climate to maximise fruit quality \cite{Smart_1985}. For example, the apple industry has been transformed in recent decades through the combination of dwarf trees, high planting densities and the use of trellis systems  \cite{Lordan,Robinson}. These innovations have allowed highly controlled canopy structures and fruit to canopy ratios, optimising the conversion of sunlight into fruit. 

In viticulture, the concept of \emph{vine balance} as being key to achieving a desired fruit quality is broadly held throughout the world's grape-growing regions. Although resistant to a specific definition the concept typically relates to the ratio between a vine's photosynthetic production and the demand for that production by the developing fruit \cite{Ravas1930,Kliewer172,Bravdo1984}. Management of the canopy through trellis systems, pruning, irrigation and leaf removal is a major determinant of vine balance, but relies strongly on the personal experience of the vineyard manager and leaf removal, in particular, is highly labour intensive.

At present, direct manipulation of grapevine canopy size and structure within a growing season is done by hand, with leaf plucking machinery (vacuum or air blast), or with cutting equipment mounted on a tractor. The former two options are amenable to adjustment as down a vine row and the latter is adjustable on a block basis. However, within-row control is difficult due to the lack of any objective measure of the canopy structure. As a result canopy adjustments tend to be relatively simple and applied over a broad area.

If objective measures of the canopy structure were available, management options would be improved, with a consequent improvement in the commercial value of the crop. By collecting this data over multiple growing seasons, a grower would be in a stronger position to target specific yield or quality outcomes \cite{edwards21digital}. Furthermore, detailed information on within-field variability would be available, highlighting under-performing or over vigorous areas and enabling automated adjustment of machinery on-the-go to minimise the effects of these.
The lack of options for growers and the limited options for researchers have resulted in a situation where the canopy structure parameters that are likely to be most useful for growers are not even known. For example, how can the role of canopy structure in determining vine productivity be defined? How should spatial and temporal changes in the light environment around the fruit best be described? 

Accurately estimating canopy structure on a per-vineyard block basis and multiple times per season poses a difficult technical challenge. The temporal fidelity requires a method that has low time burden, the requirement for accuracy and spatial detail requires a full (rather than sample based) map of the vineyard, and the temporal longevity (analysis over multiple seasons) requires an absolute and canonical data structure that is essentially unchanged as the hardware and sensors change over time \cite{edwards2020intelligent,moghadam2020digital}.

Lidar has been proven to be an effective, non-destructive scanning device for in-field canopy structure estimation when compared to the optical methods such as photogrammetry or stereoscopy \cite{dong2020semantic,kicherer2017phenoliner}. 
\cblue{There are a range of optical, non-destructive solutions available to monitor the grapevines for precision viticulture in field conditions \cite{rose2016towards}. Most of these solutions emphasise estimating phenotypic traits of grapes and berries. The biggest challenge of field deployment of optical solutions is the effect of lighting conditions on robustness and repeatability. Phenoliner~\cite{kicherer2017phenoliner} addressed changing lighting and background conditions by utilising a grape harvester as a sensing platform. However, its driving speed is limited to 0.2-0.3 $m/s$ due to limitation of the multi-view stereo 3D reconstruction.} 

Previous works have shown that lidar solutions are more reliable for estimating canopy structure, mainly due to their relative insensitivity to environmental conditions and interference caused by sun, and day/night applicability. To collect lidar data at the whole field scale, most studies mount a fixed 2D line-scan lidar \cite{moreno2020ground,siebers2018fast} or a 3D multi-beam lidar \cite{chakraborty2019evaluation,gene2019fruit} on a moving platform to scan the \cblue{crop} rows from a side view. However, due to the high-density nature of the perennial agricultural crops, many previous studies have shown these push-broom lidar settings do not provide accurate estimation of more complex canopy structure attributes such as density, porosity, and vigour \cite{murray2019novel}. \cblue{The data resulting from these push-broom lidar systems tend to be relatively sparse inside the dense canopies.}
The outer layers of the \cblue{crop} often obscure the laser path and cause shadowing effects which affects the canopy gap or density estimation. \cblue{This constraint tends to result in underestimation of the canopy density. }

For analysis of lidar data and estimating the canopy structure attributes, point clouds have been widely used as the standard 3D data format. However, point clouds lack any information about the free space in the structure. As such, it is impossible to determine from the point data alone whether a section of canopy has more points because it has a higher density of leaves, or because the lidar spent more time observing that region. Critically, the point cloud data alone does not contain information about rays that pass through the vine, therefore it is not possible to determine whether a section of canopy has a higher porosity or simply, fewer laser pulses reached that area.

\cblue{To address these limitations, we propose a novel, canopy density estimation solution using 3D ray cloud representation for perennial horticultural crops at the field scale. As opposed to the current state of the art where environments are represented using 3D point cloud maps, a key feature of our proposed solution is to utilize a ray cloud as the canonical 3D data structure for canopy attributes estimation in agriculture field conditions. } 
\cblue{Canopy density estimation is implemented using a maximum likelihood approach by integration of all ray information into a unified statistical representation, leveraging information on both lidar returns and segments of unimpeded propagation to provide an estimate of density within each voxel. The model combines data from diverse distances and angles from a moving lidar payload, and results in a distribution in each voxel. Non-homogeneous sampling density does not result in corrupted estimates, but the variability of confidence with sampling density is captured in the distribution.}

\cblue{To meet the requirement of high spatial and temporal fidelity in field conditions, we propose the application of continuous-time 3D SLAM (Simultaneous Localisation and Mapping) to a spinning lidar payload (AgScan3D) mounted on a farm vehicle. While the core SLAM algorithms have been described previously, to our knowledge this is the first time an application of continuous-time SLAM technology has been applied to agriculture settings to improve prediction of canopy structure information over extended period of time, over two growing seasons (23 months), 64 data collection campaigns resulting in a total traversal of 160 kilometres and 42.4 scanned hectares of vines. Spinning lidar provides extended fields of view, repeated lidar coverage over short time intervals and ensures that lidar beams from diverse distances and angles reach within the canopies to avoid underestimation in highly dense crops.}

\section{Related Work}
\label{sec:literature-review}

There are several previous works that review non-destructive approaches (\eg{} optical methods) for canopy structure estimation \cite{bao2019field}, but in this section, we focus on recent studies that consider lidar-based solutions under various field conditions in perennial agricultural crops, such as vineyards and orchards. 
Non-destructive lidar-based solutions have advantages over optical methods such as producing direct measurements of canopy structure and being less affected by environmental conditions such as direct sun or day/night applicability. 

Lidar-based canopy structure estimation has been widely used and studied in forestry and agriculture \cite{zhu2018improving}. 
Lidar sensor configuration for estimating vegetation indices in perennial agricultural crops, such as vineyards and orchards generally group into two main configurations. Terrestrial Laser Scanning (TLS) or Mobile Laser Scanning (MLS). Terrestrial Laser Scanning (TLS) is often stationary, mounted on a tripod to measure 3D surface of objects from one viewpoint. 
To generate a complete 3D representation of a scene, a TLS system is placed manually around objects of interest at multiple view angles. Often artificial, fixed reference targets are placed in the scene to provide spatial control points for alignment (\ie{} registration) of all 3D point clouds to generate a complete 3D representation and also avoid occlusion \cite{moorthy2011field,murray2019novel,fernandez2019estimating}.

Unlike TLS systems, Mobile Laser Scanning (MLS) systems scan scenes from a mobile platform such as ground vehicles \cite{colacco2017orange,underwood2015lidar,underwood2016mapping,chakraborty2019evaluation} or aerial vehicles \cite{torres2015high,garcia2015canopy}. 
Most of these systems use a terrestrial 2D laser scanner mounted on ground vehicles side-looking or on aerial vehicles down-looking while scanning in a push-broom setting (\ie{} trawling). In the case of ground-based MLS, these methods are often limited to one sided scanning of each vine row.  

\cblue{There are two main challenges that arise in the context of canopy structure estimation using a MLS. The first is the need for accurate registration of lidar scans. The lack of overlap between adjacent lidar scans in a push-broom MLS systems demands for a separate positioning system or prior knowledge of the environment~\cite{underwood2015lidar,underwood2016mapping}. 

A second challenge is the lack of accurate canopy structure information provided by these push-broom MLS systems in the perennial agricultural crops. Lidar beams often 
get obscured by the outer layers of the canopy which results in underestimation of structure information (sparsity) within the canopies. Due to the high-density nature of the perennial agricultural crops and lack of diverse lidar beams (distances and angles) intersecting with inside the canopies, the within parts are less observed and the canopy structure attributes such as density, porosity, and vigour are underestimated \cite{murray2019novel}. 

To meet these challenges, we propose the application of continuous-time 3D SLAM to a spinning lidar payload mounted on a mobile farm vehicle. To our knowledge this the first time a mobile, spinning lidar system has specifically been applied to address the problem of canopy structure estimation of perennial horticultural crops at the field scale. Spinning lidar provides extended fields of view, repeated lidar coverage over short time intervals and ensures that lidar beams from diverse distances and angles reach within the canopies to avoid underestimation.}

\cblue{When estimating the structure attributes of canopies, there exist numerous approaches on using registered lidar data in obtaining the crop parameters such as canopy height, width, volume, and crown diameter, volume \cite{fernandez2019estimating,murray2019novel,underwood2016mapping,chakraborty2019evaluation}. Many research methods estimate the volume of canopy as a good surrogate for describing crop structure parameters~\cite{cheein2015real,sanz2013relationship,sanz2018lidar,llorens2011ultrasonic,del2016mapping,colacco2017method}.

} 

\cblue{However, canopy volume is a poorly defined index in grapevines. The complex arrangement of shoots, leaves and tendrils is not well approximated by cuboidal sections~\cite{llorens2011ultrasonic}, alpha shapes~\cite{colacco2017method}, convex hulls or 3D grids~\cite{cheein2015real}.}
\cblue{As a result, canopy volume varies with the amount of concavity inherent in the estimate. The volume estimation is also sensitive to outliers, such as a single stray shoot. By contrast, canopy leaf area is insensitive to the positions of shoots, and the shape of the canopy.}

\cblue{In recent years, leaf area has been estimated by reconstructing individual leaves as mesh surfaces~\cite{yun2016novel}, however this requires high resolution, static or slow moving scanning device. }
\cblue{To scan whole vineyard blocks at 5-6 $km/hr$, standard lidar systems currently do not have the resolution to reliably isolate and model individual leaves. Consequently, in this paper, we estimate the leaf area density statistically per-voxel, without relying on individual leaf reconstruction.}

\cblue{Previous studies ~\cite{kamoske2019leaf,almeida2019optimizing,su2019evaluating,grau2017estimation} show that leaf area density can be estimated by counting the number of laser beams that hit within, and pass through each cubic voxel of canopy. The ratio of these counts is used to estimate the opacity of each cubic voxel, by utilising the Beer-Lambert law of energy dissipation through a turbid medium.  These approaches are computationally fast but do not not utilise the penetration depth of individual rays that contact within the volume. In \cite{hu2018estimating,chen2018estimation} the lidar point cloud is used construct a mesh model of the envelope shape of trees for which leaf density is to be estimated. Subsequently, the path length distribution is derived for rays passing through this shape, assuming they pass unimpeded. These two methods require mesh reconstruction step of an isolated crop which is not feasible in viticulture, and provide a single estimate for the entire object rather than more detailed voxel-based information.}

\cblue{Our proposed method uses penetration depth information for each individual ray on a per-voxel basis without a need for construction of mesh model. This exploits the full available data, while also accounting for varying density through the canopy (from voxel to voxel). By using a statistical estimation method (rather than reconstructing individual leaves) we are able to map whole vineyards at the typical speeds of farm utility vehicles.}

\section{\cblue{Background}}
\label{sec:background}

In this section we briefly summarize our continuous-time 3D SLAM algorithm to a spinning lidar payload, \cblue{which is utilised to generate globally registered ray clouds at the field scale}. For a more detailed description of our continuous-time SLAM please refer to our previous work \cite{bosse2009continuous,bosse2012zebedee,bosse2013place,moghadam2013line,park2018elastic,park2020elasticity}.

\subsection{SLAM}
\label{sec:slam}

The conversion of lidar range data into a globally consistent map in the absence of external sensing requires solving the Simultaneous Localisation and Mapping (SLAM) problem. We use a continuous-time SLAM framework and fuse it with GPS information in order to combine the accuracy of GPS with the precision of the lidar SLAM. 

Continuous-time SLAM is a method well-suited to spinning lidar sensors. Rather than treating the measurements in batches (or frames), it models the path of the sensor as a continuous trajectory that is receiving a sequence of lidar measurements, each at a slightly different time. 
The SLAM method utilises data from a spinning lidar and an Inertial Measurement Unit (IMU). It consists of two stages: a non-rigid registration phase, and a global registration phase. In the non-rigid registration, the candidate trajectory is optimised over a sliding window of fixed duration (in this case five seconds). The trajectory is initialised from an integration of the IMU data, and undergoes a non-linear optimisation in order to minimise the error between trajectory and IMU measurements, and between the resulting map features across multiple sweeps of the spinning lidar. 
The map features we use are surfels (surface elements); they are planar patches representing the local surface position and normal vector. The difference along the normal direction of the nearest surfels from adjacent lidar sweeps of the environment is the principle error that is minimised in the trajectory optimisation stage. 

The trajectory is modelled as a six Degree of Freedom (DoF) continuous function of time $T(\tau)=(\vec{t}(\tau),\vec{r}(\tau))$, representing a varying translation $\vec{t}(\tau)$ and rotation $\vec{r}(\tau)$. The trajectory transforms a point $\vec{p}$ at time $\tau$ from the sensor frame $S$ to the world frame $W$ as follows:
\begin{equation}
        \vec{p}_W=T_W^S(\tau)\oplus\vec{p}_S 
        =\vec{r}_W^S(\tau)\oplus \vec{p}_S+\vec{t}_W^S(\tau)
\end{equation}
where $\oplus$ is the transformation composition operator. In our case, the world frame $W$ is chosen such that the pose on commencing a scan is identity. The trajectory is optimised through a series of adjustments $\delta\vec{x}_i=(\delta\vec{t}_i, \delta\vec{r}_i)$ for $n$ iterations $i=0,\dots,n$ such that: 
\begin{equation}
       T_{i+1}(\tau)=(\delta\vec{t}_i(\tau)+\vec{t}_i(\tau), \delta\vec{r}_i(\tau)\oplus\vec{r}_i(\tau))
\end{equation}

These adjustments are found by repeatedly linearising the problem in the form $A\delta{\vec{x}}=\vec{b}$. The vector $\vec{b}$ is a stacked column of estimation errors $\vec{e}$. Each estimation error represents the difference between observation and estimation at a point along the estimated trajectory. The two main cases are IMU and surface estimation errors. 
For IMU errors, the angular error is the difference between the IMU's measured angular velocity $\vec{\omega}_{\mathrm{imu}}$ and the finite difference angular velocity of the trajectory estimation at iteration $i$ for some small time interval $\delta\tau$:
\begin{equation}
  \vec{e}_{\mathrm{imu,r}}=\vec{\omega}_{\mathrm{imu}} - (\vec{r}^{-1}(\tau-\delta\tau)\oplus\vec{r}(\tau+\delta\tau))/2\delta\tau
\end{equation}
\cblue{The IMU acceleration error is the difference between the measured acceleration $\vec{a}_{\mathrm{imu}}$ and the acceleration along the trajectory, accounting for gravity:
\begin{equation}
\vec{e}_{\mathrm{imu,a}}=\vec{r}(\tau)\oplus\vec{a}_{\mathrm{imu}} - \frac{d^2 \vec{t}(\tau)}{d\tau^2} - \vec{g}
\end{equation}}
For surface errors, surface patches are estimated using the lidar point locations $\vec{p}_W$ for the current trajectory estimate. These points are binned by time, and then binned spatially into 3D voxels. For each voxel, the points are approximated by a single \cblue{mean} location $\vec{s}_a$ \cblue{(which depends on the trajectory at the corresponding time)} and the normal of the best-fit plane $\vec{n}_a$. The surface error is then a 1D vector that measures the distance between this surface location and another surface $\vec{s}_b$ with normal $\vec{n}_b$:
\begin{equation}
    \vec{e}_{\mathrm{surf}} = (\vec{s}_b-\vec{s}_a)\bullet (\vec{n}_b + \vec{n}_a)/2 
\end{equation}
The other surface ($b$) is the nearest surface patch that was observed at a different time. 
Regardless of the choice of estimation error functions, the linear system is built in the same way. The vector $\vec{b}$ is generated as a stacked column vector of error vectors $\vec{e}$ for each point along the trajectory. The rows of $A$ are generated by differentiating each estimation error with respect to the corrections: $\frac{\partial \vec{e}}{\partial \delta\vec{x}}$. Due to the number of constraints, the linear system is overdetermined and therefore is solved in (weighted) least squares form:
\begin{equation}
       \delta\vec{x}=(A^\top WA)^{-1} A^\top W\vec{b}
\end{equation}
The diagonal weighting matrix $W$ allows a robust least squares solution by down-weighting outliers. We re-weight each constraint every iteration as a \textit{Cauchy} function of its error, which makes the algorithm an iteratively re-weighted least squares optimisation of the trajectory with respect to the IMU and lidar measurements. Once the corrections $\delta\vec{x}$ become small enough, the time window is advanced (in this case by one second), the new second of trajectory is initialised from the IMU integration and the optimisation continues on the next time window, until the whole trajectory is estimated.

This process is accurate within small time periods, but can drift over longer time periods. We account for this through the second stage, which is a global registration. This stage runs the same non-linear solver, but rather than operating on a rolling window, it considers the full trajectory, at a lower temporal resolution. This global registration is responsible for loop closure when the geometry is sufficiently close. This is because it allows nearby surfaces observed at any time over the full time interval to be aligned together. In order to guarantee coverage in the field and increase the probability of loop closure, we suggest a procedure for data collection. The data collection methodology is covered in more detail in Section~\ref{sec:experiments}.

\subsubsection{GPS}
Even with global registration, the SLAM algorithm alone is prone to error on large spatial fields like vineyards because of the aliasing problem. 
Each row looks very like its neighbouring rows, and large perturbations can cause neighbouring rows to be placed over one another. Current low cost GPS systems have sufficient accuracy to prevent the trajectory from getting close to crossing rows, however they do not have the precision in position, orientation and timing to calculate every undulation in the trajectory. We therefore fuse the low precision GPS measurements together with the lidar SLAM measurements.

This is done by adding the GPS measurements as a low weight constraint into the SLAM non-linear solver. We assume the accuracy of GPS readings to be equally distributed. The GPS location at each reading is extracted and converted into a reference coordinate frame nearby to the vineyard site, we will call it the global frame $G$. The GPS readings in this global frame are the 3D vectors $\vec{g}$. The GPS constraints are not activated until the optimiser has calculated several metres of trajectory and multiple GPS readings have been taken. At this point, the closest rigid transformation from the GPS readings to the trajectory is calculated ($H$) and the GPS readings are then applied in this trajectory's frame (which starts at identity). The error is then a 3D translation vector:
\begin{equation}
\vec{e}_{\mathrm{GPS}} = H\vec{g}-\vec{t}
\end{equation}
Before the global registration step, the entire estimated trajectory is inverse transformed by $H$, and $H$ is set to identity. The GPS constraints continue to act until the last iterations of the global registration. These final iterations assume that the surfels are well aligned and therefore prevent GPS errors from skewing the results.
With the estimated trajectory now in the globally aligned reference frame $G$, the ray cloud can be generated. The ray origins are the trajectory locations $\vec{t}(\tau)$ for every sensor reading time $\tau$, and the end points are calculated by transforming the contact points in the sensor frame, by the corresponding trajectory pose in the global frame $G$: $\vec{p}_G=T^S_G(\tau)\oplus\vec{p}_S$.

\tikzstyle{block} = [rectangle, draw, fill=blue!20, 
    text width=5em, text centered, minimum height=4em]
\tikzstyle{line} = [draw, -latex']
\tikzstyle{cloud} = [draw, ellipse,fill=red!20, node distance=3cm,
    minimum height=2em]
\tikzset{%
  cascade/.style = {%
    general shadow = {%
      shadow scale = 1,
      shadow xshift = -1ex,
      shadow yshift = 1ex,
      draw,
      thick,
      fill = white},
    general shadow = {%
      shadow scale = 1,
      shadow xshift = -.5ex,
      shadow yshift = .5ex,
      draw,
      thick,
      fill = white},
    fill = blue!20, 
    draw,
    thick,
    text width = 5em,
    minimum height = 4em,
    text centered}}    
\begin{figure}[th!]
\begin{tikzpicture}[node distance = 3.2cm, auto]
    \node [block] (slam) {SLAM};
    \node [left of=slam] (lidargps) {};
    \node [block, right of=slam] (gext) {ground extraction};
    \node [block, right of=gext] (rseg) {row segmentation};
    \node [cascade, below of=rseg](vox) {voxelise};
    \node [cascade, right of=vox] (dens) {estimate density};
    \node [block, above of=dens] (int) {spatial integration};
    \node [right of=int] (images) {};
    \node [right of=int] (graphs) {};
    \node [right of=int] (coloured_pc) {};
    \path [line] (lidargps) -- (slam) node[midway,above] {lidar, IMU} node[midway,below] {GPS};
    \path [line] (slam) -- (gext) node[midway, above]{ray} node[midway, below]{cloud};
    \path [line] (gext) -- (rseg) node[midway, above]{ray} node[midway, below]{cloud};
    \path [line] (rseg) -- (vox) node[midway, above,rotate=-90]{ray} node[midway, below,rotate=-90]{clouds};
    \path [line] (vox)  -- (dens) node[midway, above]{ray} node[midway, below]{sets};
    \path [line] (dens) -- (int) node[midway, above,rotate=90]{voxel} node[midway, below,rotate=90]{densities};
    \draw[transform canvas={yshift=0.5cm}, -latex'](int) -- (images) node[midway,below] {images};
    \path [line] (int) -- (graphs) node[midway,below] {graphs};
    \draw[transform canvas={yshift=-0.5cm}, -latex'](int) -- (coloured_pc) node[midway,below] {point cloud};
\end{tikzpicture}
\caption{An overview of the proposed canopy density estimation pipeline, from raw lidar data through to analytic results. Stacked boxes indicate multiplicity: voxelise acts on multiple vineyard rows, and density estimation acts on multiple voxel ray sets.}
\label{fig:pipeline}
\end{figure}
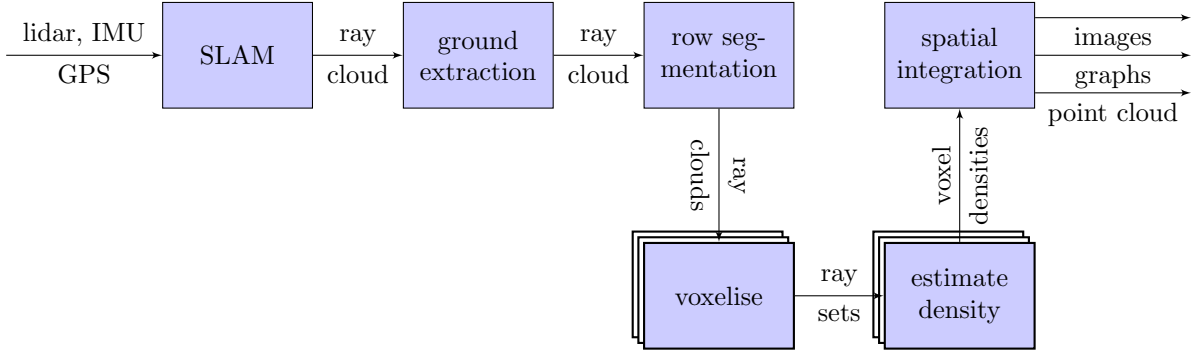

\section{Proposed method}
\label{sec:overview}

\cblue{Our proposed canopy density estimation pipeline is shown in Figure~\ref{fig:pipeline}. First, an existing continuous-time 3D SLAM algorithm is utilised to generate globally registered ray clouds at the field scale. Then, a series of automated extraction and segmentation algorithms (explained in Appendix Section) are designed and developed to remove ground and extract vineyard rows and voxelised into a set of rays per voxel. Finally, the proposed canopy density estimation (Section \ref{sec:density_estimation}) is calculated using a variable resolution method. After this we spatially integrate the densities into images, graphs and other display outputs. In the following, first we define our 3D ray cloud representation for a moving lidar system. Then, we introduce our canopy density estimation and its derivation. }

\label{sec:methodology}
\subsection{3D Ray Cloud}
Lidar maps are typically stored as the set of observed contact points in 3D space, known as a point cloud. But in order to record the structure of vines, we need to include information on the free space between contact points. We do this by recording each lidar ray (the directed line segment from sensor location to contact point) in what we call a \emph{ray cloud}.

We define a ray cloud $R$ mathematically as a set of tuples $(\vec{t}, \vec{p}, \tau)$ where $\vec{t}\in\mathbb{R}^3$ is the translation vector of the source of the ray (the 3D sensor location when the measurement was taken), $\vec{p}\in\mathbb{R}^3$ is the position where the ray contacted the environment, and $\tau\in\mathbb{R}$ is the time when the measurement was taken. 

Unlike point clouds, ray clouds also store measurements that do not receive a return signal, \cblue{due to there being no contacting surface within of the lidar's maximum range $r_{\max}$. An example are rays that point toward the sky. These rays represent a known length ($r_{\max}$) of free space, but we have no information of what is beyond that. As such, we store these in the same format as the contacting rays (with length $r_{\max}$ in the ray's direction), but flag them as having no contact point~\cite{moghadam2020structure}. }

For lidar there is a challenging problem of distinguishing non-returns due to the environment being out of range, and non-returns due to dark surfaces, glancing surface angles and points within the lidar's minimum range. In this paper we approximate this distinction by assuming that upward pointing non-returns are out of range, and downward pointing non-returns are due to one of the other cases. We discard the downward non-returns as we have no indication of their length, and the upward facing non-returns are treated as having maximum lidar range (but $\vec{p}$ is specially treated as a non-contact location in this case).  

\subsection{Canopy Density Estimation}
\label{sec:density_estimation}

Once the globally registered ray cloud is generated, a series of automated extraction and segmentation algorithms are applied to remove ground surface and segment individual rows. Each row is then divided up into voxels covering the rows of canopy in the vineyard. Each voxel is a cropped ray cloud. For more information regarding our automated ground extraction method, row segmentation and voxelisation refer to the appendices.

We then estimate canopy density $\rho_c$ within each voxel. Canopy density is the one-sided leaf area per cubic metre, in units $m^2/m^3$. 
As discussed in \cred{Section}~\ref{sec:literature-review}, existing canopy structure methods commonly estimate density using the hit count $m$ and ray count $n$ exclusively, while others have considered the distribution of distances through the volume. However, these methods ignore the penetration depth information, which adds accuracy to the estimation and is particularly useful when the number of rays is small due to high density nature of the perennial agricultural crops. In order to maximise accuracy, we include these penetration depths $x_i$ through the voxel in the formulation. \cblue{Here $x_i$ is the length of the segment of the ray at index $i$, from where it enters the voxel to where it exits the voxel (or to where it ends within the voxel).}
We estimate canopy density using the formula:
\begin{equation}
    \rho_c = g\frac{(n-1)m}{n\Sigma_{i=1}^n x_i}
    \label{eq:density}
\end{equation}
where $g$ is a constant scale factor that depends on the angular distribution of leaves relative to the rays within the voxel. Since the vineyards contain a wide distribution of leaf angles and the spinning lidar generates many ray angles, we approximate the distribution as spherical. This uniform distribution of angles has scale factor $g=2$ ~\footnote{This is the reciprocal of the G parameter in classic Leaf Area Index models based on the Beer-Lambert law for light attenuation in uniform mediums~\cite{yan2019review}.}.

\cred{
Equation~\eqref{eq:density} can be written as:
\begin{equation}
    \rho_c = g\hat{\lambda}d(n),
    \label{eq:factored}
\end{equation}}
where \cred{$\hat{\lambda}$ is the leaf density estimate that we find in Section \ref{ss:DensityEstimate}}:
\cred{\begin{equation}
\hat{\lambda} = \frac{m}{\Sigma_{i=1}^n x_i}
\label{eq:leafdensity}
\end{equation}}
The parameter $\lambda$ represents the density of infinitesimal leaves, which are modeled as an unbounded homogeneous Poisson point process.
The final term is a heuristic debiasing factor:
\begin{equation}
d(n)=\frac{n-1}{n}
\label{eq:debias_func}
\end{equation}%
This is an approximate scale factor, to account for more realistic scenarios than the simple leaf model that \cred{$\hat{\lambda}$} uses. Specifically, it accounts for leaf distributions that are bounded within a voxel, and for leaves that are finite in size. The following section provides a derivation of these formulae, together with numerical validation.

\subsection{Canopy Density Derivation}
\label{ss:DensityEstimate}
Our underlying model of the canopy is as a turbid medium; we then include a debiasing function to compensate for elements lacking from this model. A turbid medium models the leaves as infinitesimal and spatially distributed with uniform probability. The leaf positions within a voxel are modeled as a homogeneous Poisson point process. This means that the probability of an intersection occurring in a short interval $\delta x$ is $\lambda\delta x$, where $\lambda$ is the leaf density in units of intersections per ray length. Consequently, the probability density function of a lidar range measurement $x\geq 0$ given $\lambda$ is:
\begin{equation}
f(x|\lambda) = \lambda\exp\{-\lambda x\} 
\end{equation}

When we measure the medium with $n$ lidar rays, we obtain a series of path lengths $y_1,\dots,y_n$ which the ray would have been able to pass through the voxel if unimpeded by leaves, as well as the set of indices $\mathcal{I}\subseteq\{1,\dots,n\}$ of the paths that intersected a leaf. Finally we include the penetration depths $x_1,\dots,x_n$ which are the length through the voxel to interception when $i\in\mathcal{I}$ and equals the unimpeded length $y_i$ when $i\not\in\mathcal{I}$. The probability that a given ray $i$ is intercepted is then:
\begin{equation}
P(i\in\mathcal{I}) = \int_0^{y_i} \lambda\exp\{-\lambda x\}dx = 1-\exp\{-\lambda y_i\}
\end{equation}
Conversely, the probability that a ray will not be intercepted is:
\begin{equation}
P(i\notin\mathcal{I}) = \int_{y_i}^\infty \lambda\exp\{-\lambda x\}dx = \exp\{-\lambda y_i\}
\end{equation}
The probability density function of a ray \cblue{conditioned on it being intercepted in the region} $0\leq x_i\leq y_i$ \cblue{is given by the unconditional pdf divided by the probability that it is intercepted within that region:}
\begin{equation}
f(x_i|\lambda;y_i) = \begin{cases}
\frac{\lambda\exp\{-\lambda x_i\}}{1-\exp\{-\lambda y_i\}}, & 0 \leq x_i \leq y_i \\
0, & \mbox{otherwise}
\end{cases}
\end{equation}

The likelihood of our observation is then:
\begin{align}
l&(\mathcal{I},x_{\mathcal{I}}|\lambda) = p(\mathcal{I}|\lambda) \prod_{i\in\mathcal{I}}f(x_i|\lambda;y_i) \\
&= \prod_{i\notin\mathcal{I}} \exp\{-\lambda y_i\} \prod_{i\in\mathcal{I}} 
(1-\exp\{-\lambda y_i\})\left(\frac{\lambda\exp\{-\lambda x_i\}}{1-\exp\{-\lambda y_i\}}\right) \\
&= \lambda^m \exp\left\{-\lambda\sum_{i=1}^n x_i\right\}
\label{eq:likelihood}
\end{align}
where $m=|\mathcal{I}|$. This can be interpreted as the likelihood of observing $m$ events in a total penetration depth of $\sum_{i=1}^n x_i$. This formulation is closely related to estimation with censored samples \cite{Coh91}, since rays which exit a voxel without intersecting leaves have effectively been censored, i.e., we are unable to observe the eventual intersection that would have occurred if the ray had continued through an infinitely sized voxel with uniform density.

The natural conjugate prior for the exponential distribution is the Gamma distribution, which has density~\cite{BDA3}: 
\begin{equation}
\textrm{Gam}(\lambda;\alpha,\beta) = \frac{\beta^\alpha}{\Gamma(\alpha)}\lambda^{\alpha-1}\exp\{-\beta\lambda\}
\label{eq:prior}
\end{equation}

Given this prior distribution~\eqref{eq:prior} and the likelihood function~\eqref{eq:likelihood}, the posterior distribution is known to be the Gamma function: 
\begin{equation}
\textrm{Gam}\left(\lambda;\alpha + m,\beta + \sum_{i=1}^n x_i\right) 
\end{equation}

We do not assume prior information in this paper, therefore we commence from $\alpha=\beta=0$. The \cred{conditional} mean, mode and variance of this Gamma distribution are then given by:
\begin{equation}
\textrm{E}[\lambda] = \frac{m}{\sum_{i=1}^n x_i}; \quad 
\textrm{Mode}[\lambda] = \max\left\{\frac{m-1}{\sum_{i=1}^n x_i},0\right\}; \quad
\textrm{Var}[\lambda] = \frac{m}{(\sum_{i=1}^n x_i)^2}
\label{eq:BayesEstimates}
\end{equation}
Thus the Gamma distribution provides a straight-forward method for estimating the leaf density parameter $\lambda$ as well as its uncertainty. 

\subsubsection{Estimator bias}
The previous section would suggest an estimator of:
\begin{equation}
\hat{\lambda} = \frac{m}{\sum_{i=1}^n x_i}
\end{equation}
However, this can be shown to be biased, i.e., $\textrm{E}[\hat{\lambda}]\neq \lambda$. Estimators of $\lambda^{-1}$ are examined with censoring in \cite{Coh91}, but the quantity of interest in this case is the leaf area per unit volume, $\lambda$. Without censoring (i.e., $y_i=\infty$ and $\mathcal{I}=\{1,\dots,n\}$, $m=n$), the variable $\sum_{i=1}^n x_i$ is Gamma distributed as $\textrm{Gam}\left(n,\lambda \right)$, such that its reciprocal is inverse Gamma, with expected value $\lambda/(n-1)$, and therefore 
\begin{equation}
\hat{\lambda}=(n-1)/\sum_{i=1}^n x_i
\label{eq:uncensored}
\end{equation}
is an unbiased estimator, with standard deviation $\sigma_{\hat{\lambda}}=\frac{\lambda}{\sqrt{n-2}}$. 

With censoring, no such result exists, but in Section \ref{sec:expturbid} we experimentally demonstrate that the heuristic debiasing factor in Eq~\eqref{eq:debias_func} exhibits sufficient accuracy for typical observations.

\section{Experiments}
\label{sec:experiments}
Demonstrating the validity of our method is challenging due to the lack of a practical means of obtaining ground truth of canopy density at vineyard block scales. It is impractical to de-leaf multiple vines and compare that to our calculated densities for multiple time points during a growing season.

Instead, we assess our method's validity through a sequence of experiments, which together provide a weight of evidence that the theoretical method is effective, and transfers as expected when applied in the field. 
In this section, simulation and field experiments are presented to provide qualitative and quantitative analysis.

\subsection{Simulation}
\label{sec:sim}

Simulation is an important first step in testing new estimation algorithms. It allows a ground truth, and very large sample sizes. It also allows the theory to be validated without interference from unmodelled components, such as woody material and sensor noise. 

We present a series of numerical simulations of increasing realism. Firstly, a numerical test is developed to demonstrate the accuracy of Eq~\eqref{eq:density} under the turbid medium model on a one-dimensional voxel, and to validate the debiasing factor. Secondly, a Monte Carlo simulation is carried to show the accuracy of Eq~\eqref{eq:density} for finite sized triangular leaves in a 3D voxel. Finally, a Monte Carlo simulation is used to show the accuracy improvement in using a spinning lidar compared to a push-broom, trawling lidar.  

\subsubsection{Simulation of Turbid Medium in 1D}
\label{sec:expturbid}

\begin{figure}[t]
\centering
\begin{subfigure}{.44\linewidth}
\centering
\includegraphics[width=0.85\linewidth]{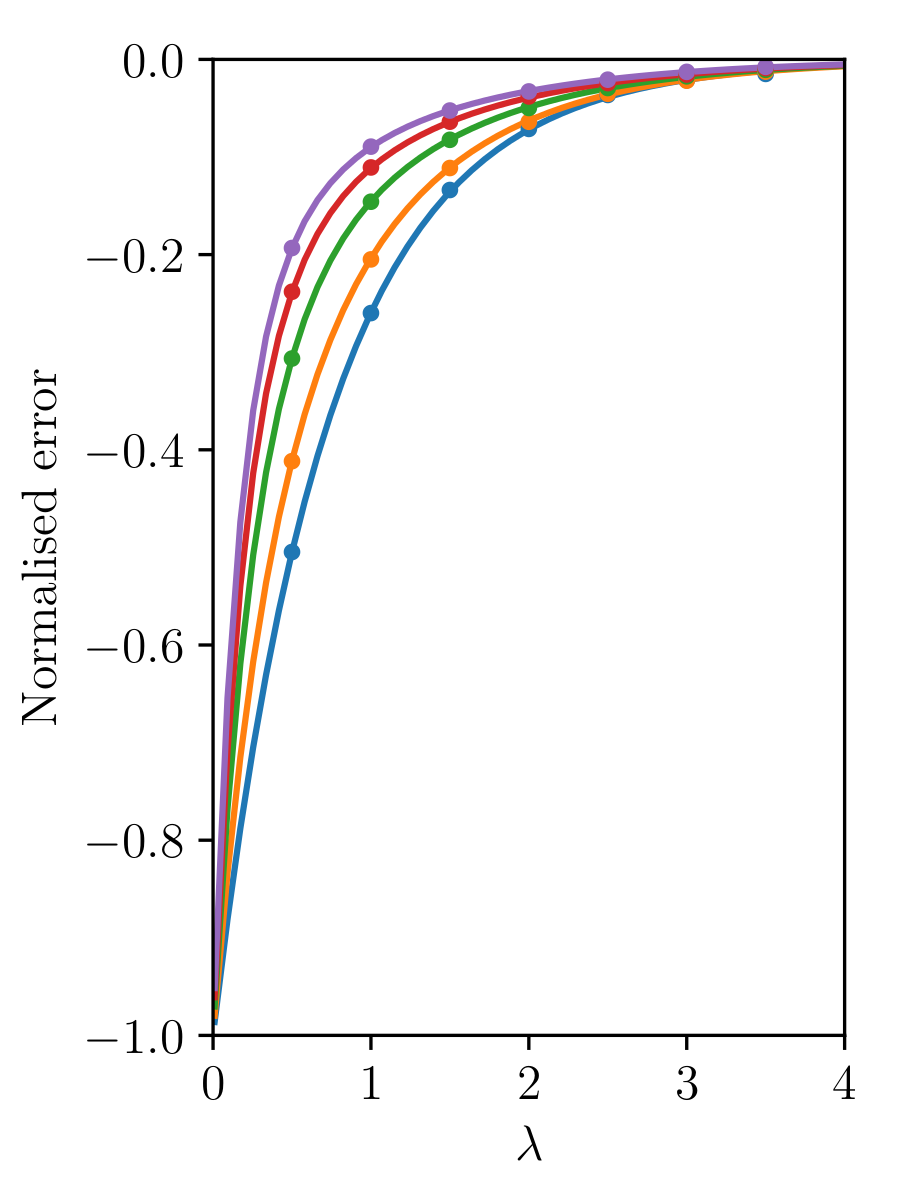}
\caption{}
\end{subfigure}%
\begin{subfigure}{.44\linewidth}
\centering
\includegraphics[width=0.85\linewidth]{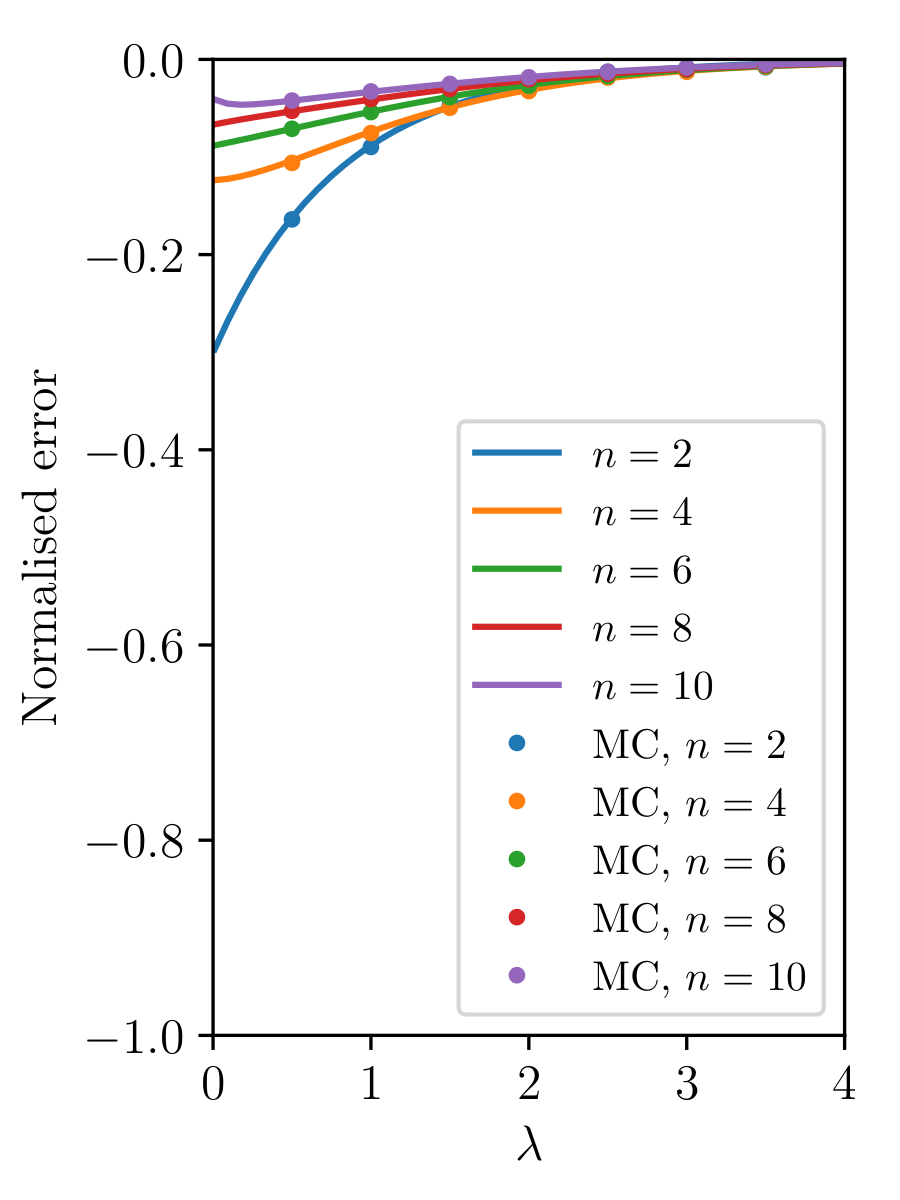}
\caption{}
\end{subfigure}
\caption{\cred{Normalised density estimation error $(\hat{\lambda}-\lambda)/\lambda$ in estimate $\hat{\lambda}$ relative to true density $\lambda$, over a range of $n$, $\lambda$, showing an underestimation bias. (a) shows result for maximum likelihood estimator, while (b) shows result for our proposed debiased estimator.}}
\label{fig:EstBias}
\end{figure}

Firstly, in Figure~\ref{fig:EstBias}\cred{(a)}, we evaluate the estimator Mode[$\lambda$] from Eq~\eqref{eq:BayesEstimates}, \cred{which in the absence of prior data coincides with the maximum likelihood estimator,} relative to its true value $\lambda$, by plotting the \cred{normalised error $(\hat{\lambda}_{ML}-\lambda)/\lambda$} for a range of values of $\lambda$ and for multiple numbers of rays $n$. A value of zero indicates an unbiased estimate. The dots are evaluated using $10^6$ Monte Carlo simulations per point, while the lines are the result of numerical integration based on the theoretical model. Throughout, the voxel depth is assumed to be a constant value $y_i=1\;\forall\;i$, and therefore $x_i\leq 1\;\forall\;i$.

Secondly, in Figure~\ref{fig:EstBias}\cred{(b)}, we evaluate our debiased estimator \cred{$\hat{\lambda}_D = d(n)\hat{\lambda}$} from Eq~\eqref{eq:leafdensity} and~\eqref{eq:debias_func} relative to the true $\lambda$ in the same manner, in order to demonstrate the improved accuracy of our debiased density estimator. These plots show that the debiased estimator of Eq~\eqref{eq:density} is more accurate than the standard maximum likelihood estimator, particularly when $\lambda$ is small.

\subsubsection{Simulation of Triangular Leaves in 3D Voxel}
\begin{figure}[t]
\centering
\begin{subfigure}{.44\linewidth}
\centering
\includegraphics[width=0.85\linewidth]{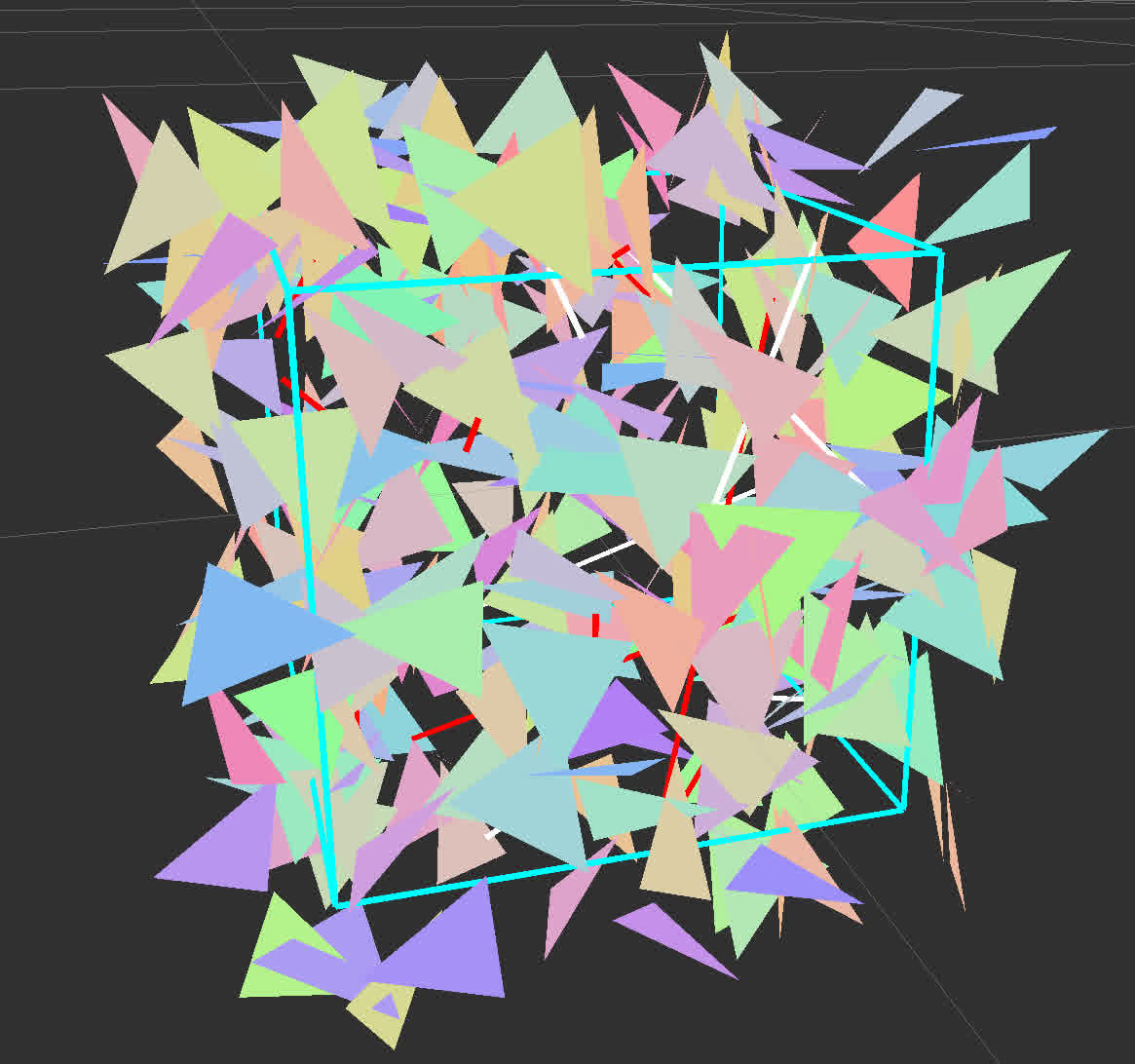}
\caption{}
\end{subfigure}%
\begin{subfigure}{.44\linewidth}
\centering
\includegraphics[width=0.85\linewidth]{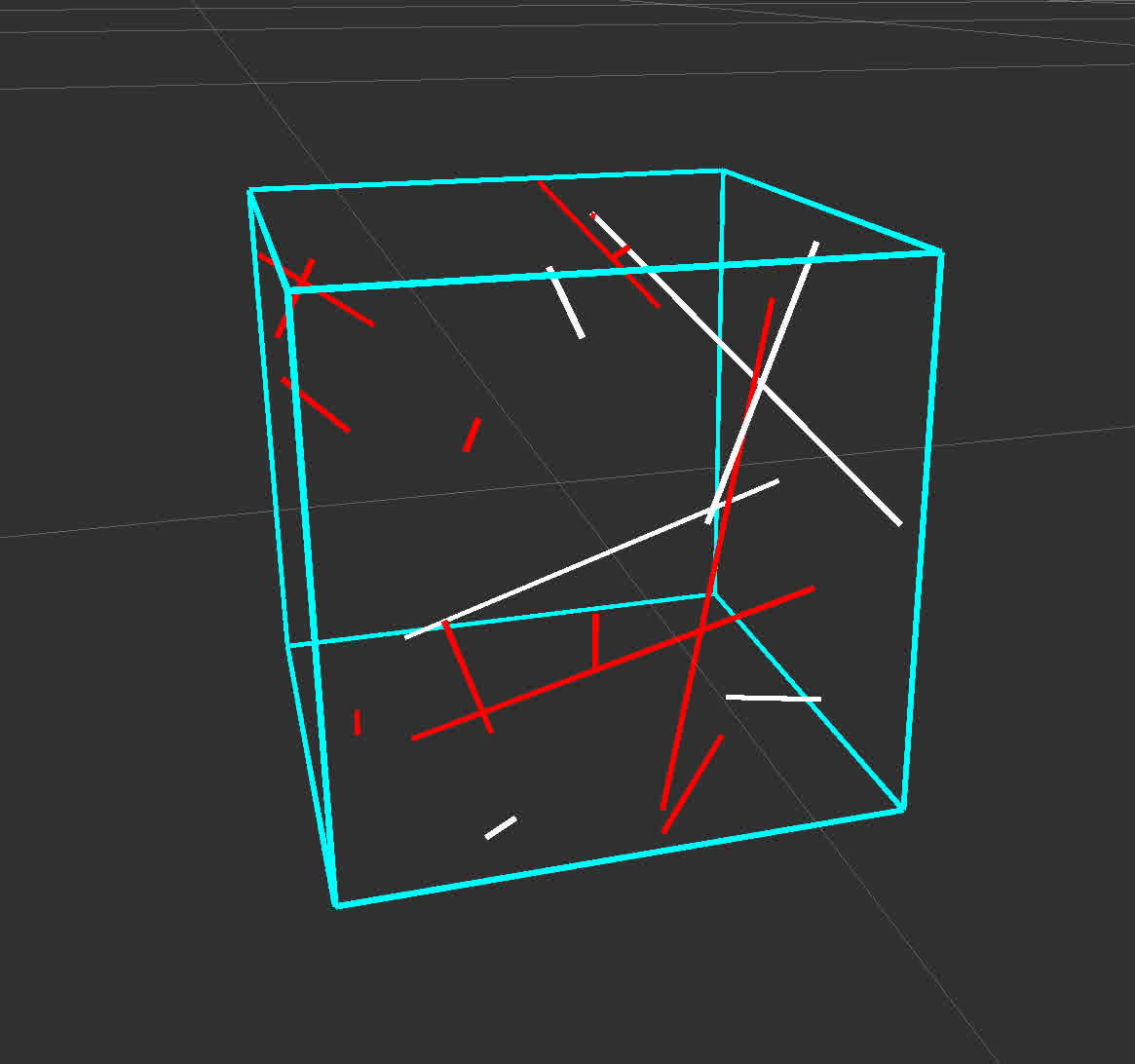}
\caption{}
\end{subfigure}
\caption{(a) Visualisation of randomly distributed triangular leaves of side length 2.5 cm, and total leaf area density of 30 $m^2/m^3$ within the 10 cm wide voxel (cyan). Red rays contact a triangle, white rays pass through. (b) with triangles removed for better visibility. }
\label{fig:triangle_leaves}
\end{figure}

Next we validate the model under a more realistic scenario. We first consider arbitrarily oriented rays passing through a 3D voxel. Secondly, we use finite leaf size rather than infinitesimal leaf sizes. 
We simulate the ray intersections through a single 3D voxel, with equilateral triangles of constant size representing the leaves. There is a fixed probability for the presence of a leaf in any unit of volume within the voxel, it therefore replaces the uniform density medium with a uniform density probability distribution of fixed size leaves over the voxel. The leaves are sampled from a uniform distribution of orientations, as seen in Figure~\ref{fig:triangle_leaves}.

We use a 10 cm voxel size and leaves of side length 2.5 cm to 10 cm, with canopy densities from 1 $m^2/m^3$ to 40 $m^2/m^3$, which roughly corresponds to between 3 and 40 leaves in a voxel. These values are selected based on real vineyard settings, and our spinning lidar parameters.

For each choice of leaf side length $l$ and total leaf area $A$ we calculate the expected number of leaves per cubic metre as $w^{-3}\frac{A}{\sqrt{3/4}l^2}$ where the denominator is the area of each triangular leaf, and $w$ is the voxel width. 
We then generate a randomly placed set of leaves within the voxel according to this density, and record the actual canopy density from the total area $\bar{A}$ of the triangles that reside within the voxel: $\rho_c=\bar{A}/w^3$.

We then ray trace $n$ randomly positioned and oriented rays, through the voxel, to generate the set of penetration depths $x_1,\dots,x_n$, and record the number of intersections $m$. We calculate the error between the estimated density (without debiasing) and the ground-truth density as $e=gE[\lambda]-\rho_c$.
Running this over $N$ independent trials allows us to estimate the error between this finite leaf size model and the continuous model. We obtain the normalised error: $\hat{e}=\frac{\langle e\rangle}{\langle \rho_c \rangle}$ where $\langle\rangle$ denotes the average value over all $N$ trials. This allows us to evaluate any bias in this more realistic setting.

\begin{figure*}[t!]
\centering
\begin{subfigure}[b]{.45\textwidth}
  \centering
  \includegraphics[width=.99\linewidth]{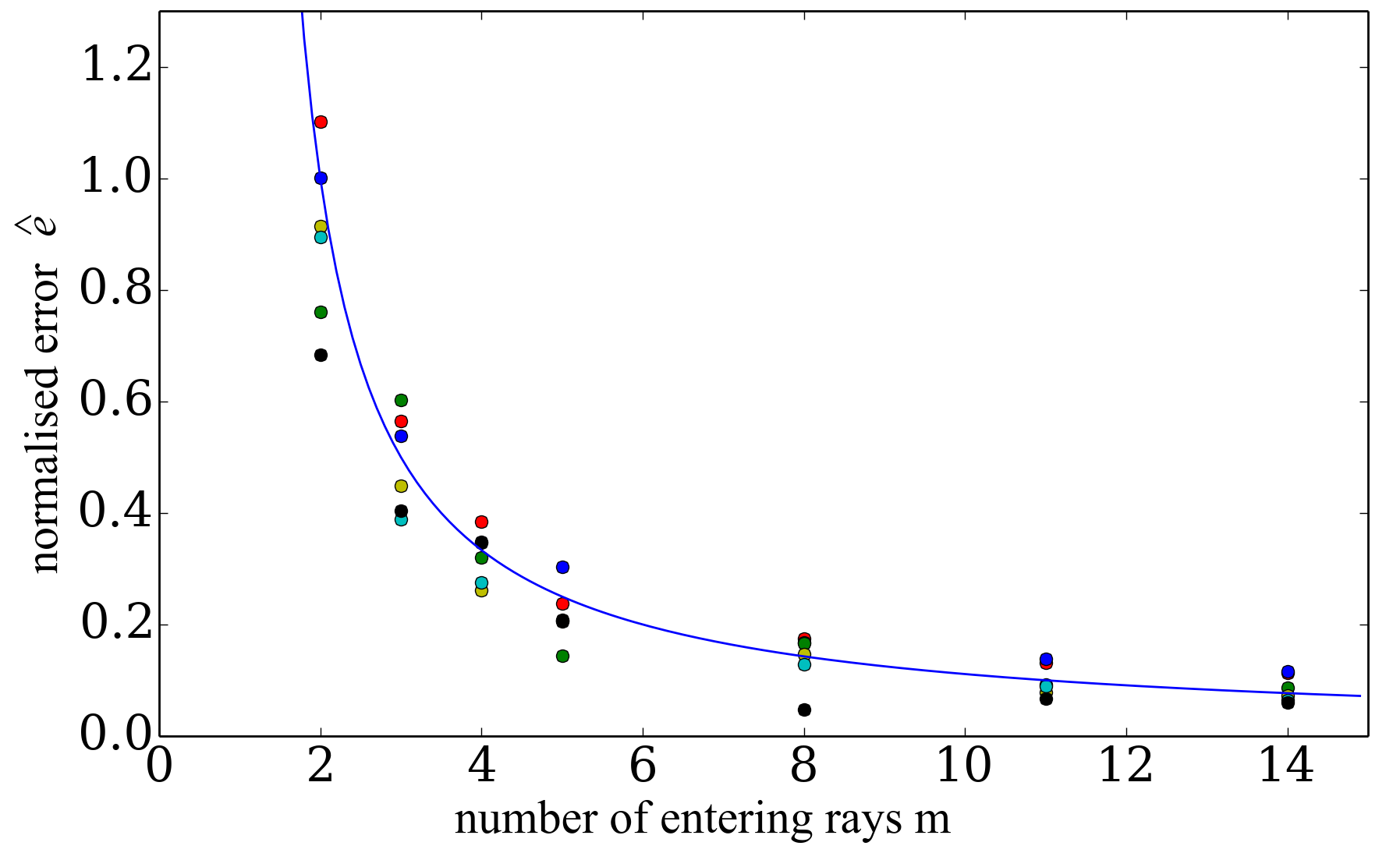}
  \caption{}
\end{subfigure}%
\begin{subfigure}[b]{.50\textwidth}
  \centering
  \includegraphics[width=.99\linewidth]{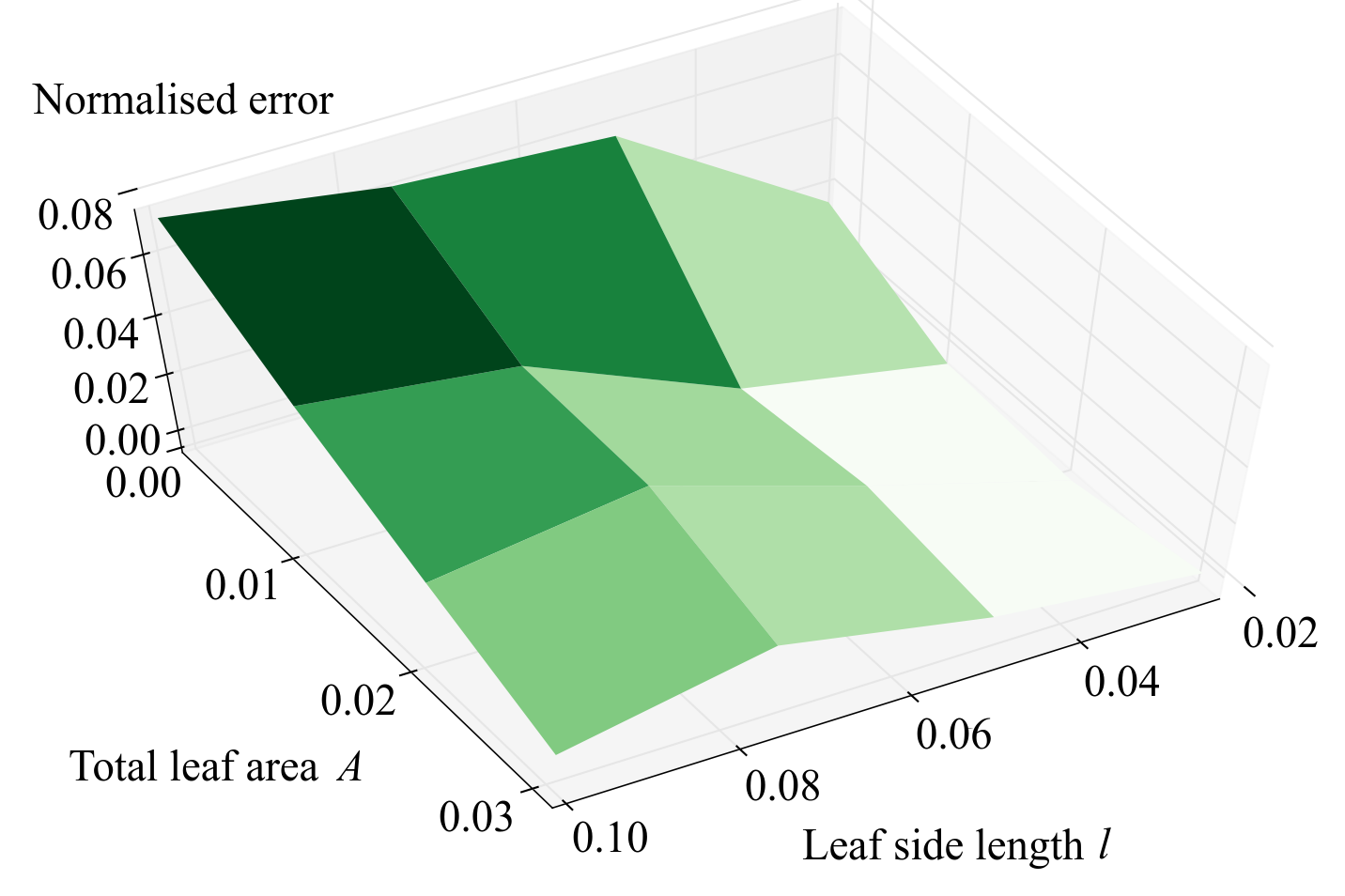}
  \caption{}
\end{subfigure}
\caption{(a) Normalised error $\hat{e}$ for six ($l$,$A$) configurations, with respect to the number of entering rays $m$. Overlaid is expected bias function $1/(m-1)$ if Eq~\eqref{eq:debias_func} were accurate. The six configurations are red:(0.1,0.012),  blue:(0.1,0.04), green:(0.05,0.003), yellow:(0.05, 0.01), cyan:(0.025, 0.012) and black:(0.025, 0.001). (b) Normalised error for $m=20$ entering rays with $N=1600$ trials, after debiasing. Leaf side length $l=$2.5 cm to 10 cm, leaf area $A=$ 0.001 to 0.031 $m^2$. }
\label{fig:bias_graph}
\end{figure*}

We then plot the normalised error obtained for six different combinations of leaf side length and canopy density, under 4000 trials where the number of entering rays $n$ varies from 2 to 14. The results are shown in Fig~\ref{fig:bias_graph}~(a) and are plotted against the expected value from the debiasing function (Eq~\eqref{eq:debias_func}), to verify that accuracy of this debiasing function. 

The results show that Eq~\eqref{eq:debias_func} is well matched to the bias in $E[\lambda]$, over a large parameter space. It provides evidence that the debiasing function is approximately the correct shape with respect to the number of rays and including this debiasing term improves our error estimation under the randomly distributed leaf model. 

The second experiment includes the debiasing function and uses it to plot the normalised error in leaf area for differing leaf sizes and canopy densities. We use $m=20$ entering rays with $N=1600$ trials, with leaf side length $l=$ 2.5 cm to 10 cm and leaf area $A=$ 0.001 to 0.031 $m^2$. The results are plotted in Fig~\ref{fig:bias_graph}~(b).

The graph shows a normalised error of no more than 0.08 (or 8\%) over a large range of leaf sizes and total leaf areas. The average error is between 3\% and 4\%. This result verifies that our proposed density estimation is sufficiently unbiased for estimating canopy density for volumes containing realistic sized leaves. The low gradient of the surface indicates that our turbid medium model transfers to finite sized leaves with little additional error. 

\subsubsection{Comparison of Trawling and Spinning lidar in Simulation}
\begin{figure*}[t]
\centering
\begin{subfigure}[b]{.33\textwidth}
  \centering
  \includegraphics[width=.99\linewidth]{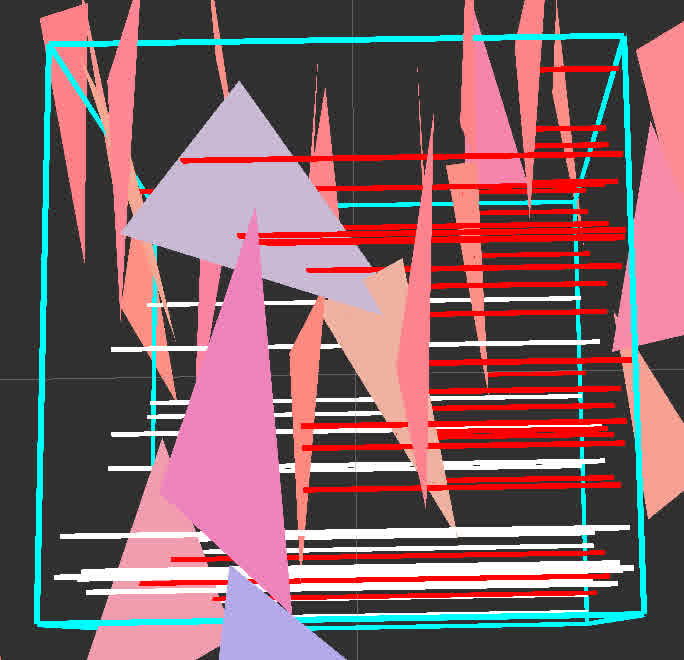}
  \caption{}
\end{subfigure}%
\begin{subfigure}[b]{.33\textwidth}
  \centering
  \includegraphics[width=.99\linewidth]{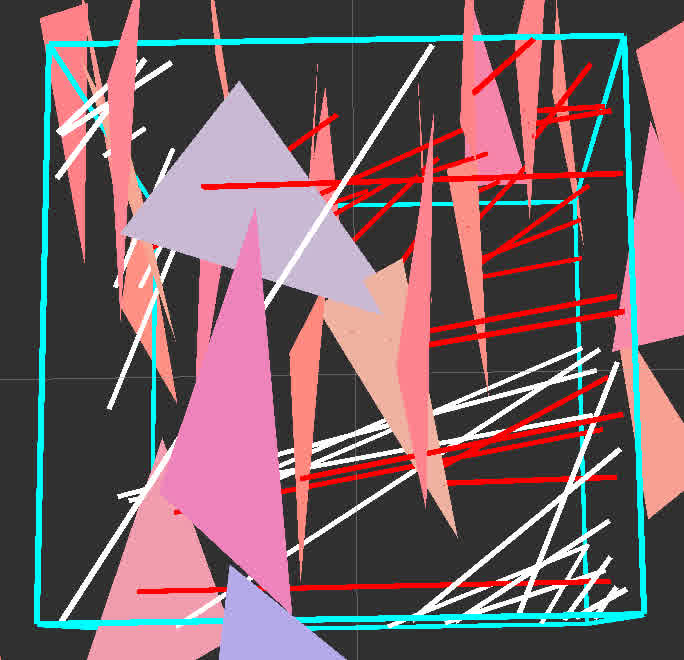}
  \caption{}
\end{subfigure}
\caption{The simulated ellipsoidal leaf distribution with normals biased to the x axis (horizontal in image), eccentricity 10,1,1. (a) rays from a trawling lidar are aligned on the x axis. (b) rays from the spinning lidar are horizontal and distributed 0 to 70 degrees from the x axis. Red rays have collided with a leaf.}
\label{fig:spinning}
\end{figure*}

Next, we compare the accuracy of canopy density estimation for push-broom (trawling) lidar, compared to spinning lidar systems. 
We run a Monte Carlo simulation using 400 trials of a distribution of triangular leaves within a 10 cm voxel, with a total leaf area of 0.02 $m^2$, using 50 rays. We model the trawling lidar rays as aligned along the x axis, while spinning lidar rays have a varying angle with height, within any one voxel the rays are approximately parallel. We model the spinning lidar as a horizontal distribution of rays of angle 0 (along the axis) up to 70 degrees, which is approximately the distribution of rays covered by the spinning lidar. See Fig~\ref{fig:spinning}. 

For each lidar we calculate the percentage error in canopy density estimate for four different distributions of leaf normal vectors. These are elliptical distributions that represent leaves that approximately face towards the x axis, y axis, z axis respectively, and a fourth that is a spherical distribution. The results are shown in Table~\ref{tab:one}.

\begin{table}[b!]
\centering
 \begin{tabular}{|c c c|} 
 \hline
 leaf normal eccentricity & trawling estimation error & spinning estimation error \\ [0.5ex] 
 x,y,z &  & \\ [0.5ex] 
 \hline
 \hline
 10,1,1 & 67.9\% & 45.5\%\\ 
 1,10,1 & -44.6\% & -1.1\%\\ 
 1,1,10 & -68.5\% & -68.5\%\\ 
 1,1,1 & 3.7\% & 2.6\%\\ 
 \hline
\end{tabular}
\caption{Percentage error between estimated and actual leaf area for simulated triangle leaves of 6cm side length in a 10 cm voxel, using 50 rays and 400 trials. Shown for four leaf normal distributions, where each distribution is sampled from an ellipsoid of the given dimensions. Trawling rays are along the x axis, and the z axis is vertical.}    
\label{tab:one}
\end{table}

The results indicate that a spinning lidar provides better density estimates than a fixed, trawling lidar system over a wide range of leaf distributions. 
It is only in the horizontal leaf case that the percentage error for fixed and trawling lidar are similar. For vineyards, where leaves fall in many different angles \cite{Mabrouk1997}, the use of a spinning lidar makes the spherical leaf distribution (relative to lidar ray direction) more appropriate than with a trawling lidar. This reduces the requirement to provide an empirical model of leaf angle distribution, which can be location and season dependent, and challenging to quantify and verify.

Together, this set of numerical results indicates that our per-voxel density estimation function (Eq~\eqref{eq:density}) and spinning lidar setup are well positioned to estimate canopy density in vineyards with large or small leaves. Next, we need to demonstrate that these results transfer into field experiments. 

\subsection{Field Experiments}
\label{sec:expreal}

In this section, we outline our field experiments to evaluate the generalization, robustness and repeatability of our proposed method. Field experiments were conducted at four vineyard sites in South Australia, which varied in vineyard structure and vine management. Data were collected over two growing seasons (23 months), resulting in a total traversal of 160 kilometres and 42.4 scanned hectares of vines. 

Firstly, we describe AgScan3D, a mobile 3D spinning lidar system used for our field experiments, then we provide details of datasets collected at multiple time points over two seasons. Finally, we provide a series of specific experiments to further evaluate our proposed method. 

The underlying index that we estimate is the canopy density $\rho_c$, which is the one-sided leaf area per cubic metre, in $m^2/m^3$. We present this data in a number of reduced forms by spatially integrating over one or more axes. Two-dimensional density images are generated by integrating over the single unrendered axis, to give the one-sided leaf area per area represented in the image, in $m^2/m^2$. The leaf area density along each row is calculated by spatially integrating over both axes perpendicular to the row, to give a one-sided leaf area per metre, in $m^2/m$. We will refer to both of these reduced indices as `integrated canopy density', with integration axes being evident from the image and the given units. The reason that we use integrated, rather than average,  canopy density, is that the extents of the canopy are not clearly defined, and cannot be estimated accurately. We therefore ensure that we only use indices that are suited to accurate measurement. Furthermore, due to the way the vine (or tree) grows in response to light availability, the cross section will not have a uniform density and thus the two dimensional variation is likely to be of more value to the grower.

\subsubsection{AgScan3D: mobile 3D spinning lidar system}
\label{sec:hardware}

\begin{figure}[t!]
\centering
\begin{subfigure}{.3\linewidth}
\centering
\includegraphics[width=0.99\linewidth]{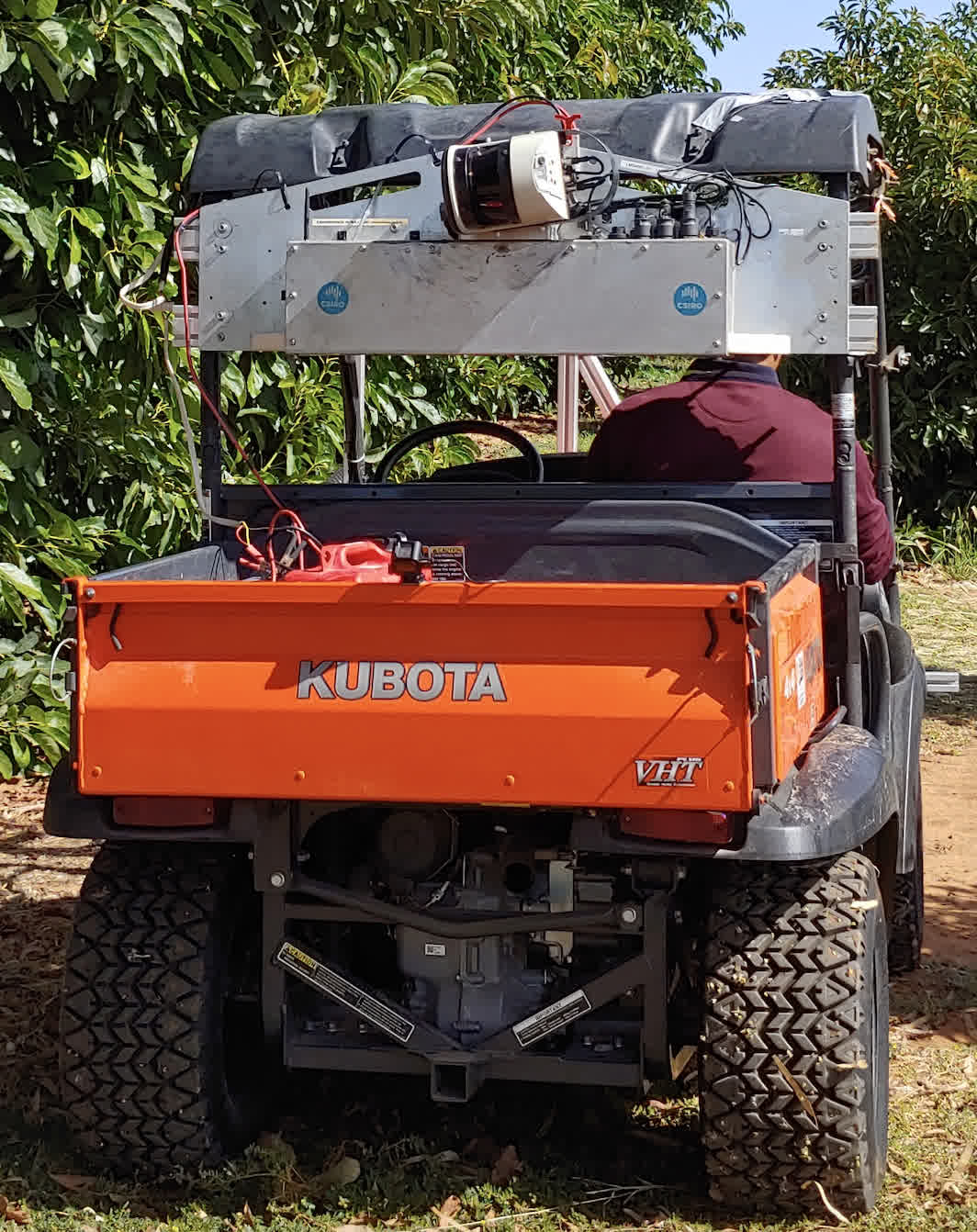}
\caption{}
\end{subfigure}%
\begin{subfigure}{.7\linewidth}
\centering
\includegraphics[width=0.95\linewidth]{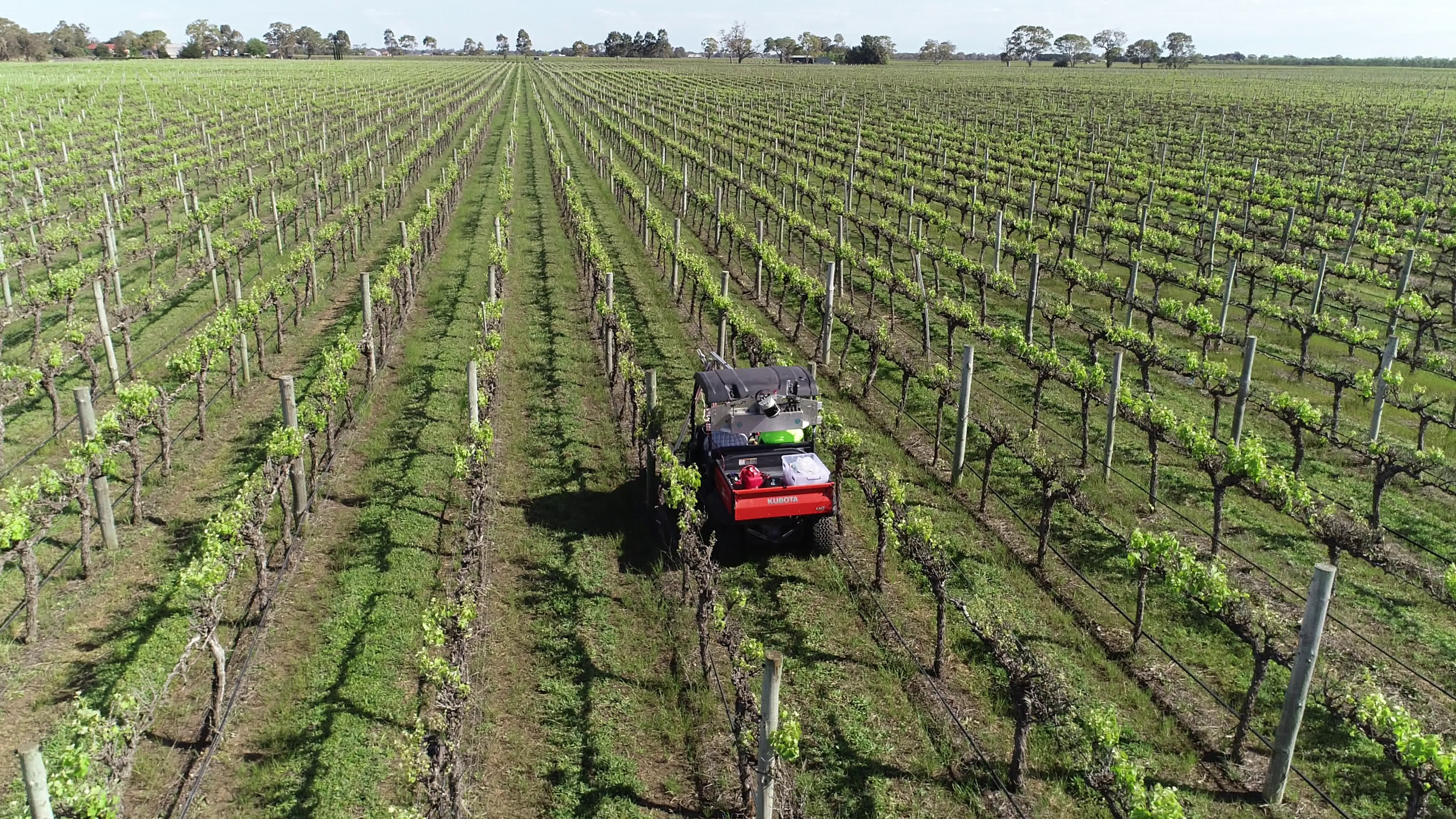}
\caption{}
\end{subfigure}
\caption{(a) AgScan3D: mobile 3D spinning lidar system retrofitted to the back of Kubota farm vehicle (b) A photograph taken from an unmanned aerial vehicles showing the AgScan3D scanning the Katnook vineyard.}
\label{fig:AgScan3D}
\end{figure}

AgScan3D sensing payload is comprised of a 3D spinning lidar, a Microstrain 3DM-GX3 IMU, a GPS unit and an onboard embedded computer. The 3D spinning lidar is comprised of a commercial 2D Laser range finder (SICK LMS511-20100) which has been mounted on a spinning platform (Figure~\ref{fig:AgScan3D}). The platform spins the lidar at a rate of 0.5 Hz. 

The laser beams have a divergence of 0.66 degrees, a frequency of 100 Hz and maximum range of 40 meters. The frequency of measurements is 10 Hz for the GPS unit, and 100 Hz for the IMU. The IMU is mounted in a fixed frame at the centre scan line of the lidar sensor. The GPS unit is mounted on the top of the vehicle. Intrinsic and extrinsic calibration between sensors is calculated beforehand. We used the Robot Operation System (ROS) for data acquisition and synchronisation on the Linux Ubuntu server 16.04 embedded machine. Users access the system via a tablet UI for starting, stopping and real-time diagnosis of the system during scans.  
The AgScan3D sensing payload is then retrofitted to an existing farm utility vehicle. As shown in Figure~\ref{fig:AgScan3D}, the device is mounted in the back of the vehicle for convenience and to avoid obstruction of view for the driver. However, the SLAM algorithm and canopy density estimation algorithm are not sensitive to forward or rear facing setup. AgScan3D uses SLAM for estimating accurate 6 DoF pose of the vehicle in order to process the lidar data into a globally registered 3D ray cloud.  

\subsubsection{Dataset} 
\label{sec:dataset}

We captured datasets at commercial vineyards in two wine grape growing regions of South Australia, Mclaren Vale and Coonawarra. The exact number of scans per growing season varied between seasons and sites, but covered the period of primary growth and key phenological periods. For example, pre-budburst, pre-flowering, maximum canopy growth (typically veraison) and post-harvest. All of the commercial vineyards scanned used mechanical harvesting, which resulted in the loss of a significant amount of leaves. Further testing was undertaken at the University of Adelaide's experimental vineyard at their Waite Campus in Adelaide and at the South Australian Research and Development Institute's (SARDI) experimental vineyard at Nuriootpa in the Barossa Valley. These represented a variety of vineyard structure, with row width's from 2 to 3.2 m, vine spacing from 1.5 to 2.4 m and cordon/wire heights from 0.95 to 1.4 m. Management styles ranged from a dense ``sprawl'', where vines canes were allowed to hang down from the cordon/stem to highly managed Vertical Shoot Position (VSP), where canes were held upright. The degree of within season canopy and irrigation management also varied, affecting the dynamic size and density of the vine canopies. Each capture started with a stationary period of approximately 10 seconds, to allow the IMU biases to be calibrated. The vehicle was then driven down between the vine rows at approximately 5-6 km/h. Rows were not usually scanned in order, due to the tight turning radius required, but every row in the block or section of block was scanned, with a single scan per row.

Scanning a block took 30-45 minutes depending on the row length and number of rows, and was straightforward as there was no burden on the user other than driving the vehicle. Our field dataset comprises of 64 scans, across 4 vineyards over a period spanning 23 months, with a combined total of approximately 93000 vines, a total of 42.4 scanned hectares, representing 160 km of vineyard rows and approximately 33 hours of scanning time. Further information is given in Table~\ref{table:totalscans}.

\begin{table}[t]
\caption{Summary of data captured with our system. Between February 2018 and end Dec 2019 (23 months, 2 seasons). Coverage based on a row spacing of 2.75, 2, 3 and 3 metres for the four vineyards respectively.}
\begin{tabularx}{1.0\textwidth} { 
  | >{\raggedright\arraybackslash}X | >{\raggedright\arraybackslash}X | >{\raggedright\arraybackslash}X | >{\raggedright\arraybackslash}X | >{\raggedright\arraybackslash}X | >{\raggedright\arraybackslash}X | 
  >{\raggedright\arraybackslash}X |}
  \hline
  Vineyards & No. of Scans & Unique Rows Scanned & Total Vines & Total Distance & Total Coverage & Total Duration \\
  \hline
  Accolade, Mclaren Vale & 36 & 39 & 49700 & 89.5 km & 24.6 ha & 18.5 hours \\
  \hline
  Katnook Estate, Coonawarra 
  & 8 & 12 & 22900 & 34.4 km & 6.9 ha & 6.5 hours \\
  \hline
  Rymill Wine, Coonawarra 
  & 17 & 12 & 18500 & 33.3 km & 10.0 ha & 6.8 hours \\
  \hline
  SARDI Nuriootpa, Barossa Valley 
  & 3 & 10 & 2000 & 3.0 km & 0.9 ha & 49 minutes \\
  \hline\hline
  \textbf{Total} & \textbf{64} & \textbf{73} & \textbf{93100} & \textbf{160 km} & \textbf{42.4 ha} & \textbf{32.6 minutes} \\
  \hline
\end{tabularx}
\label{table:totalscans}
\end{table}

\subsubsection{Effectiveness Experiment}

\begin{figure*}[t!]
\centering
\begin{subfigure}[b]{0.3\textwidth}
  \centering
  \includegraphics[width=.99\linewidth]{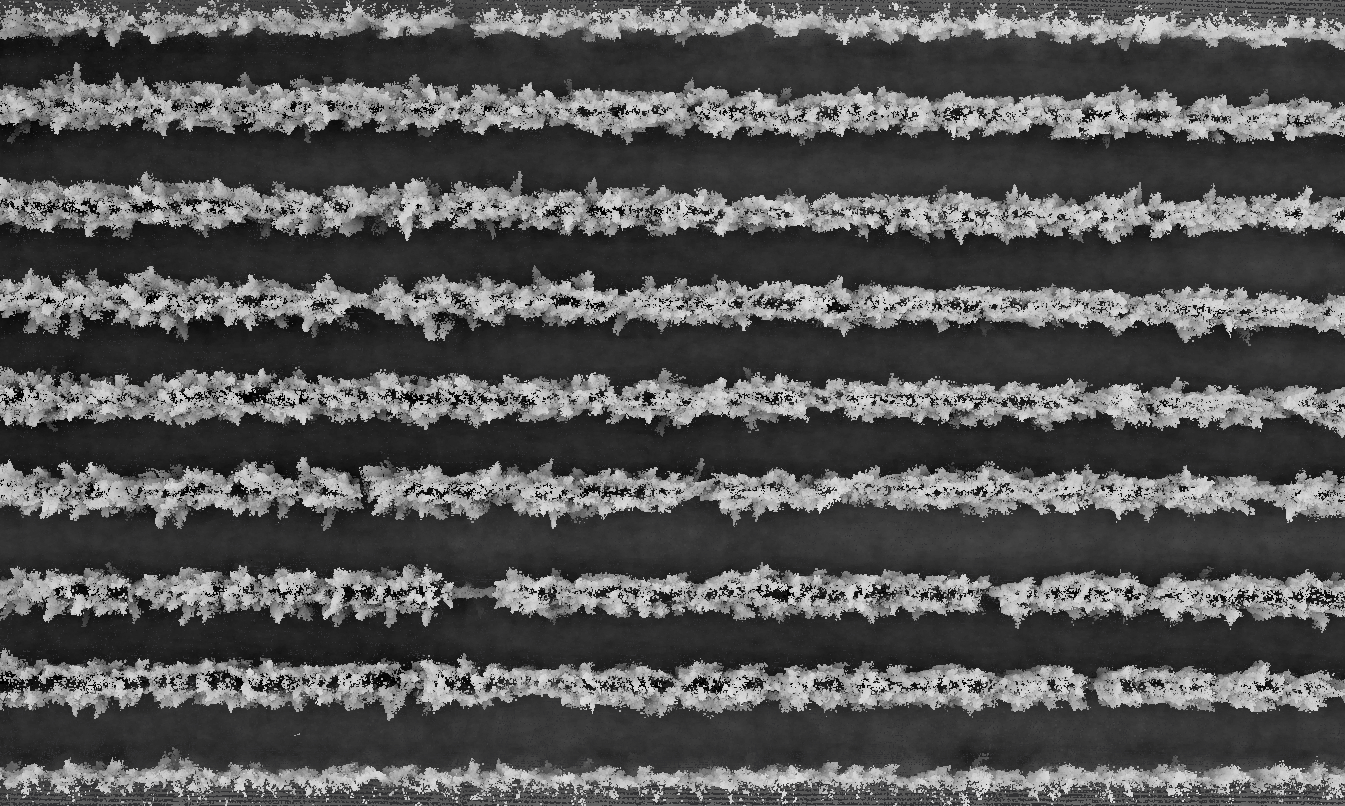}
\end{subfigure}
\begin{subfigure}[b]{0.3\textwidth}
  \centering
  \includegraphics[width=.99\linewidth]{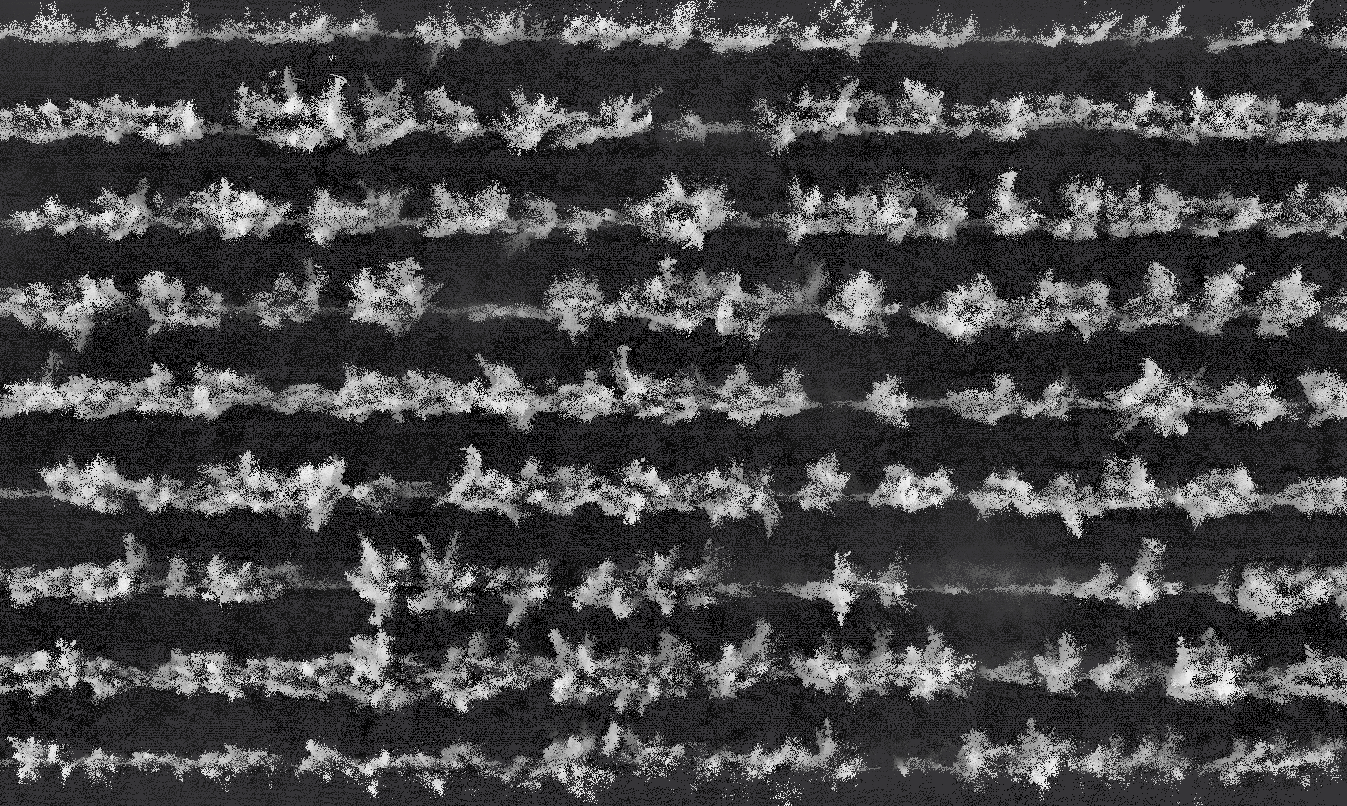}
\end{subfigure}
\begin{subfigure}[b]{0.3\textwidth}
  \centering
  \includegraphics[width=.99\linewidth]{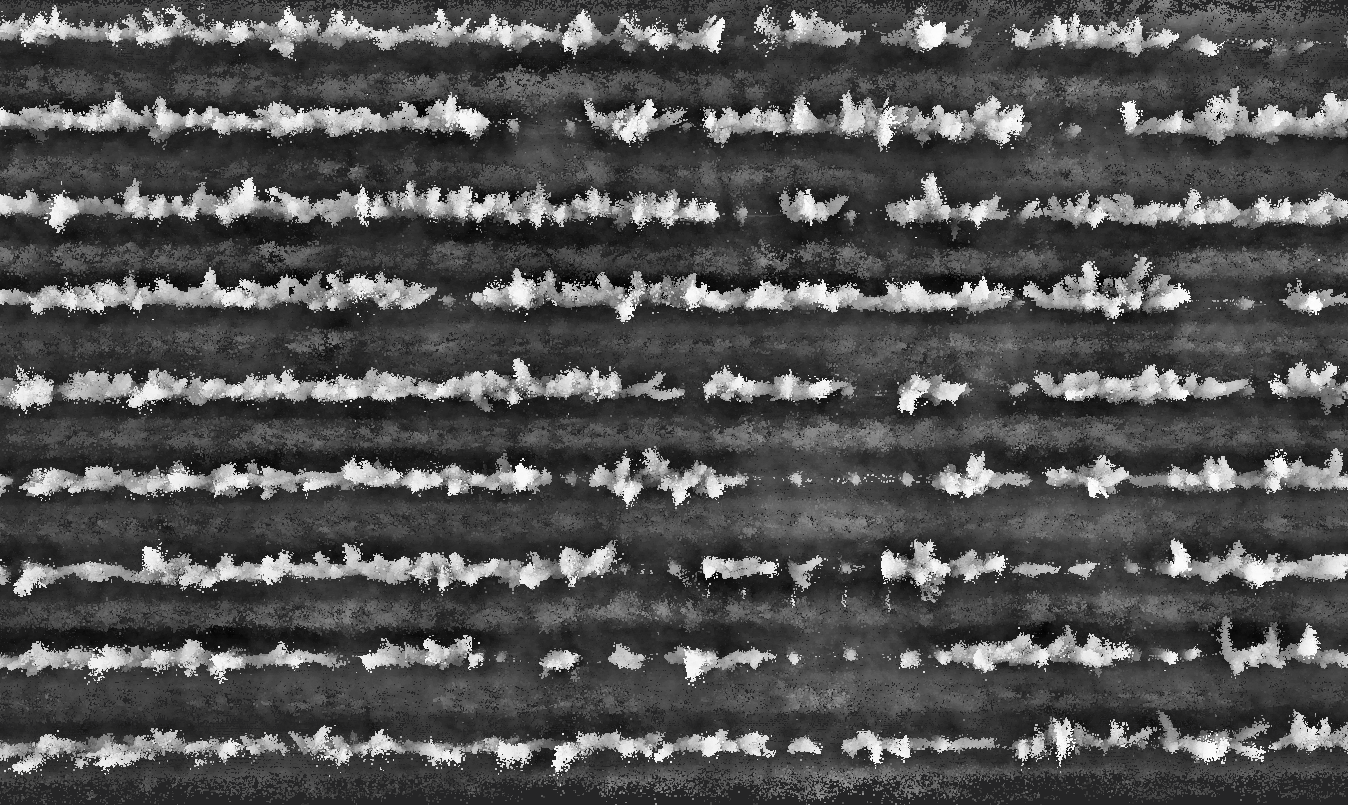}
\end{subfigure}
\begin{subfigure}[b]{0.3\textwidth}
  \centering
  \includegraphics[width=.99\linewidth,trim={0 0.645cm 0 1.19cm},clip]{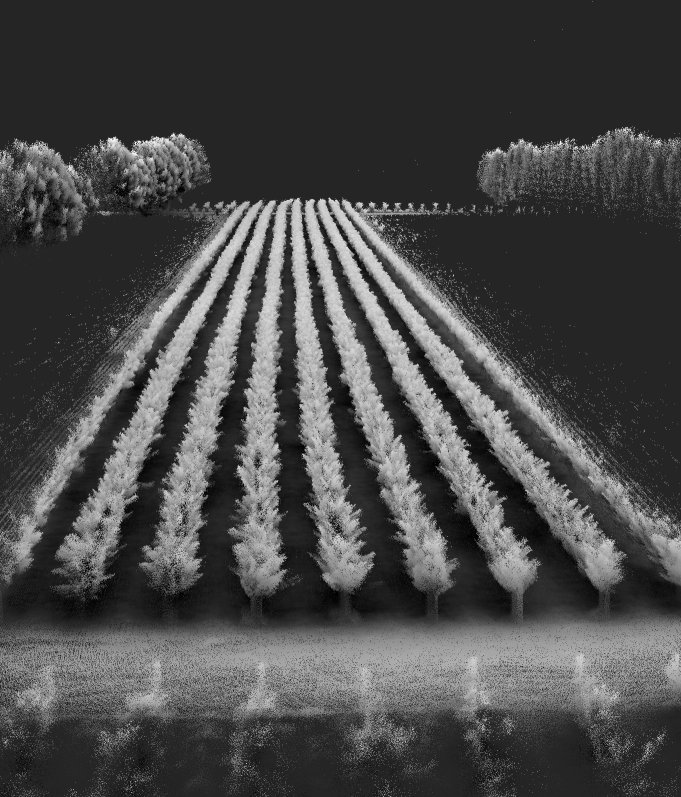}
\end{subfigure}
\begin{subfigure}[b]{0.3\textwidth}
  \centering
  \includegraphics[width=.99\linewidth,trim={0 2.68cm 0 1.6cm},clip]{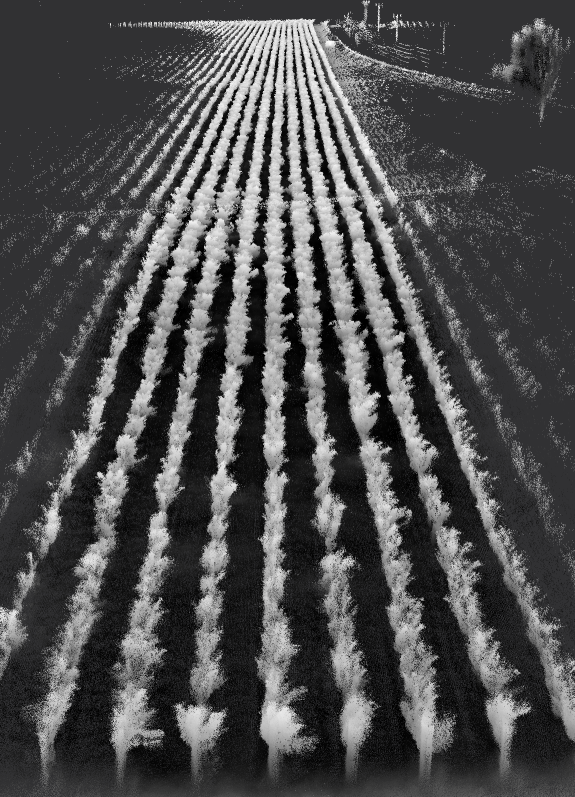}
\end{subfigure}
\begin{subfigure}[b]{0.3\textwidth}
  \centering
  \includegraphics[width=.99\linewidth]{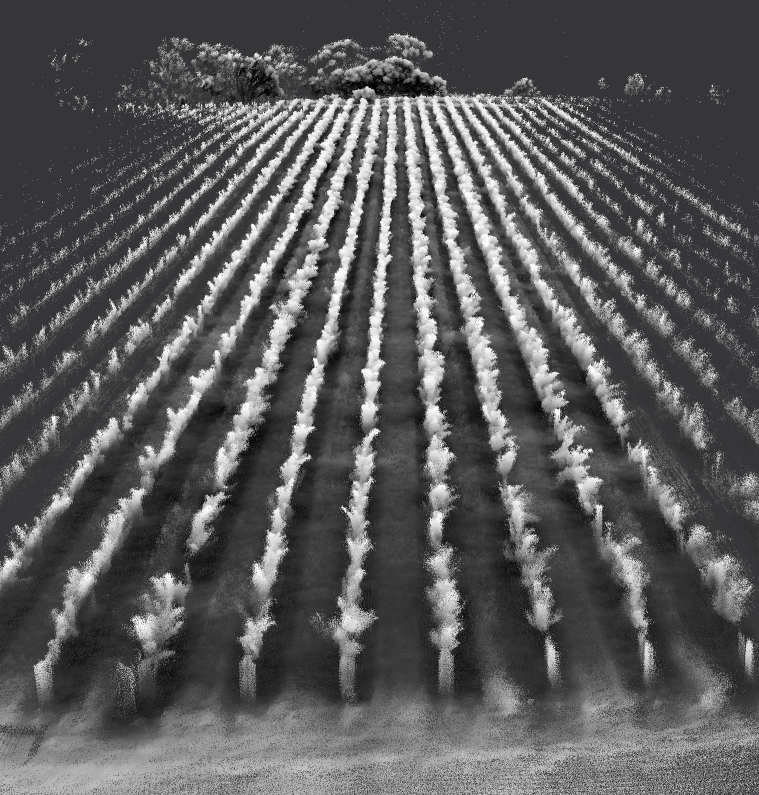}
\end{subfigure}
\caption{Globally registered point cloud from our AgScan3D: mobile 3D Spinning lidar system represented three varieties of vineyard structure. from left to right: Rymill, Katnook and Mclarenvale vineyards.}
\label{fig:pointcloud}
\end{figure*}

We demonstrate the generalization and effectiveness of our SLAM solution to generated globally consistent maps in Figure~\ref{fig:pointcloud} where we show qualitative results over three varieties of vineyard structure. Figure~\ref{fig:pointcloud} demonstrates globally consistent point clouds generated from our SLAM solution using the mobile 3D spinning lidar system (AgScan3D). We only show point clouds for visualisation purposes. As it is outlined in Section~\ref{sec:methodology}, the canopy density estimation uses 3D ray clouds as the data source, not point clouds. 

\begin{figure*}[t!]
\centering
\begin{subfigure}[b]{0.4\textwidth}
  \centering
  \includegraphics[width=.99\linewidth]{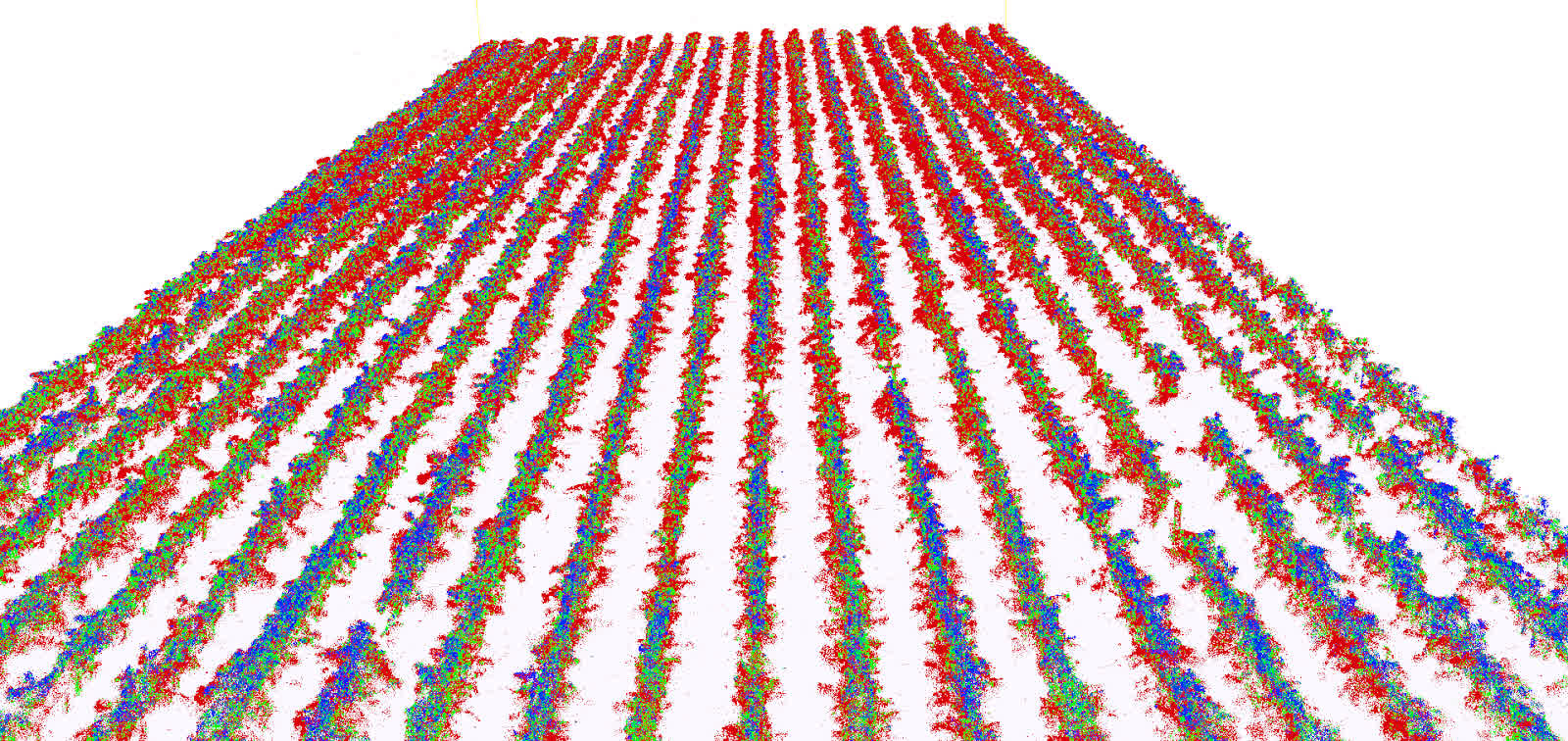}
  \caption{}
\end{subfigure}
\begin{subfigure}[b]{0.4\textwidth}
  \centering
  \includegraphics[width=.99\linewidth,trim={0 2.5cm 0 0},clip]{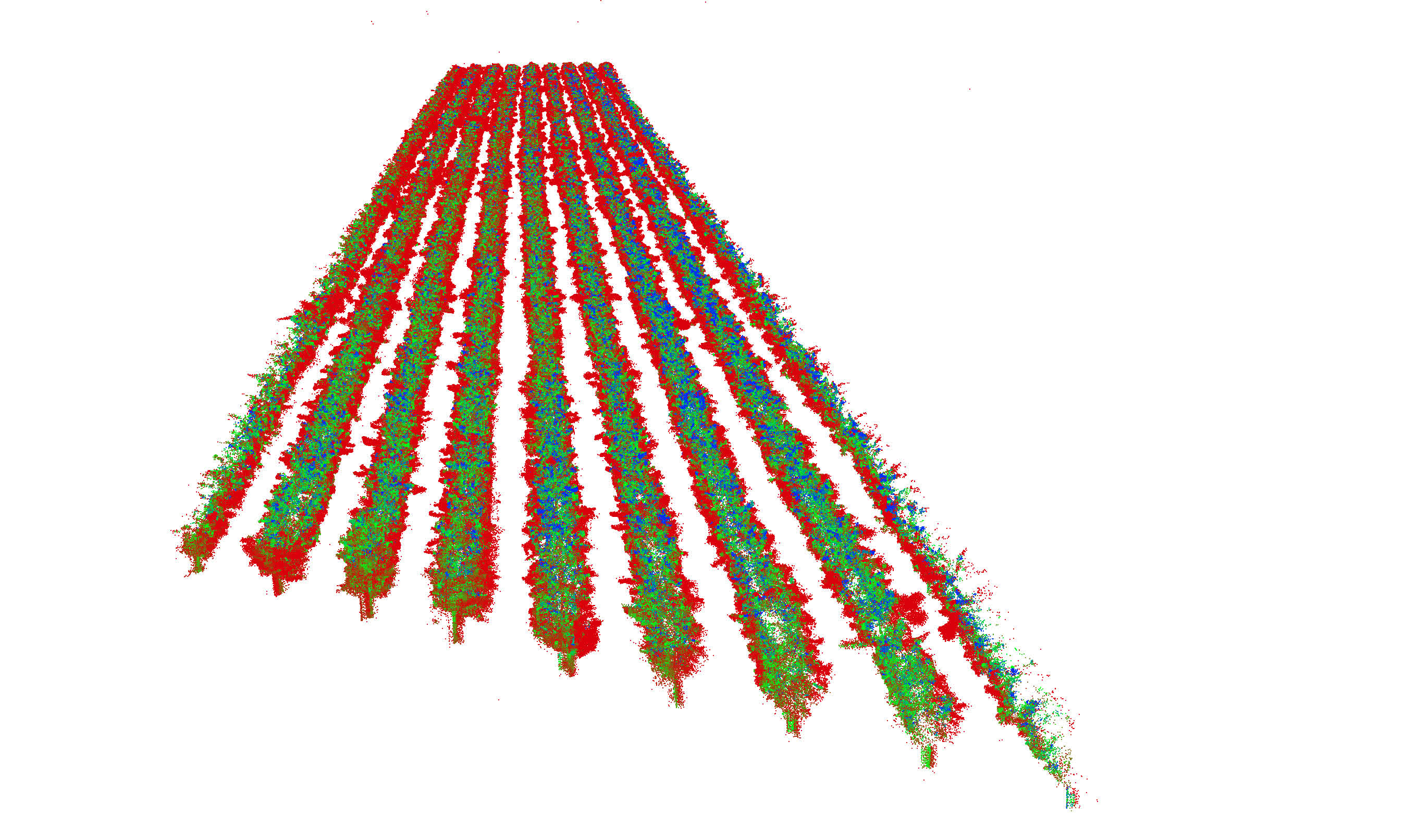}
  \caption{}
\end{subfigure}
\begin{subfigure}[b]{0.16\textwidth}
  \centering
  \includegraphics[width=.99\linewidth]{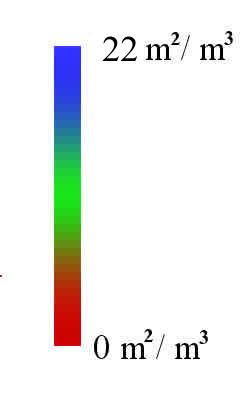}
\end{subfigure}
\caption{Globally registered point cloud false colorised according to the canopy density ($m^2/m^3$) of the voxel they reside in, using a red-green-blue colouring. (a) Mclarenvale 18/03/2019 (b) Rymill 28/02/2019.}
\label{fig:canopydensityfield}
\end{figure*}

Next, Figure~\ref{fig:canopydensityfield} demonstrates two globally registered point clouds from Mclarenvale and Rymill vineyards, false colorised by the canopy density ($m^2/m^3$) of the voxel they reside in using a red-green-blue colorbar. Colorbar is to help associate pixel values with the canopy densities.   

\begin{figure*}[t!]
\centering
\begin{subfigure}[b]{0.75\textwidth}
  \centering
  \includegraphics[width=.99\linewidth]{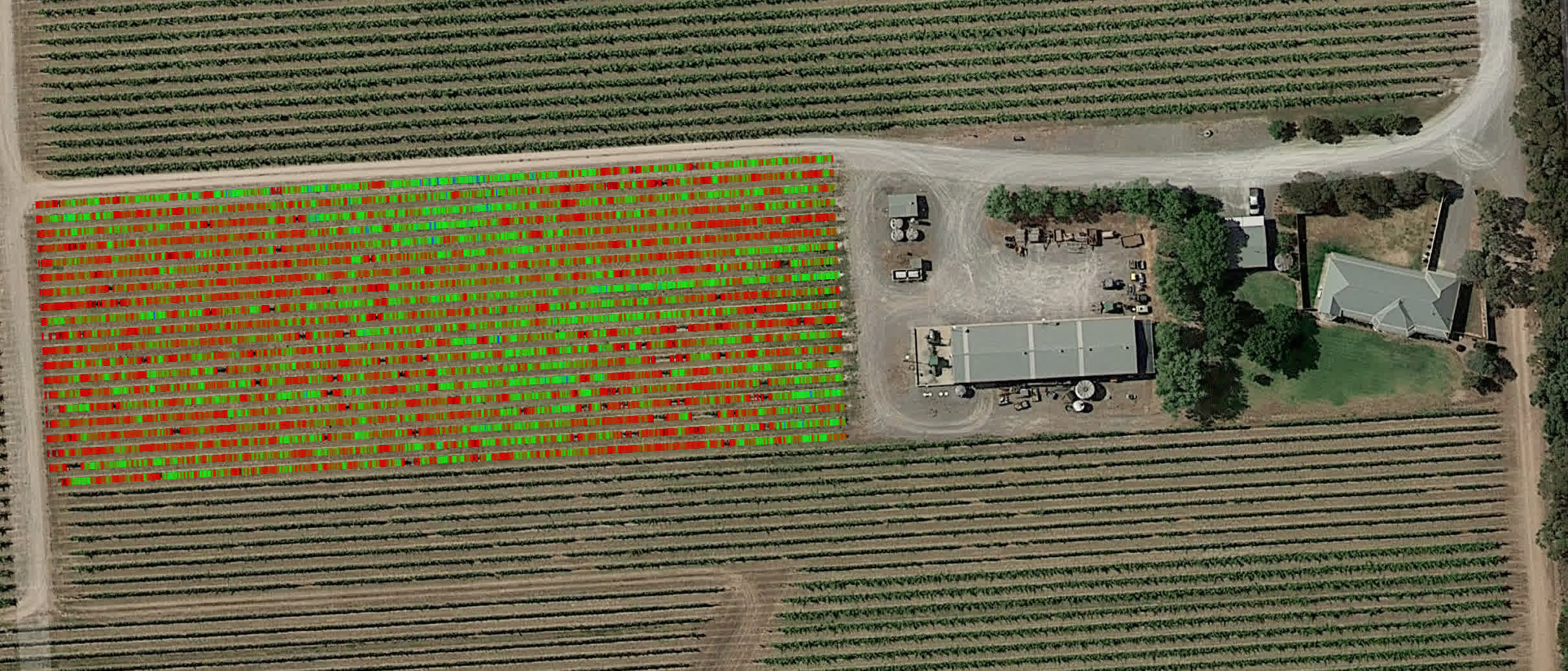}
\end{subfigure}
\begin{subfigure}[b]{0.11\textwidth}
  \centering
  \includegraphics[width=.99\linewidth]{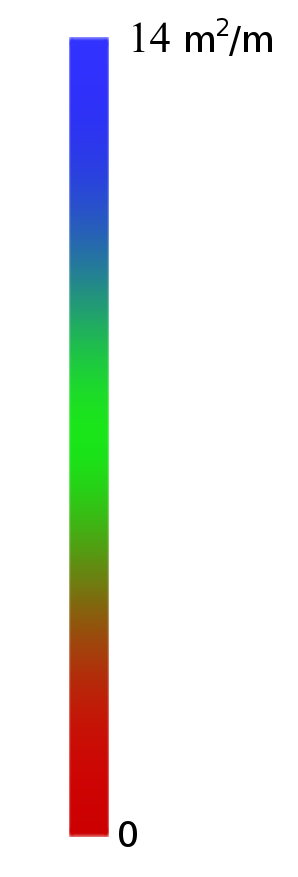}
\end{subfigure}
\caption{Integrated canopy density per metre along each row rendered over the overhead satellite view of the Mclarenvale vineyard 11/02/19 . The same red-green-blue density gradient is used as in previous images.} 
\label{fig:aerial}
\end{figure*}

In addition, the estimated densities were integrated per metre along each row and overlaid on top of the overhead satellite view of the Mclarenvale vineyard as shown in Figure~\ref{fig:aerial}. 

\subsubsection{Robustness Experiment}

\begin{figure*}[t]
\centering
\begin{subfigure}[b]{0.90\textwidth}
  \centering
  \includegraphics[width=.99\linewidth]{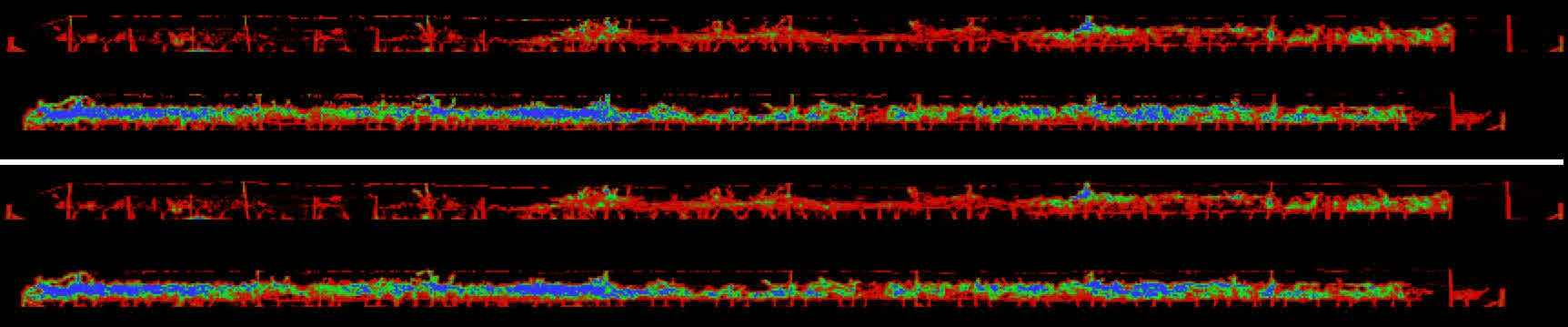}
\end{subfigure}
\begin{subfigure}[b]{0.066\textwidth}
  \centering
  \includegraphics[width=.99\linewidth]{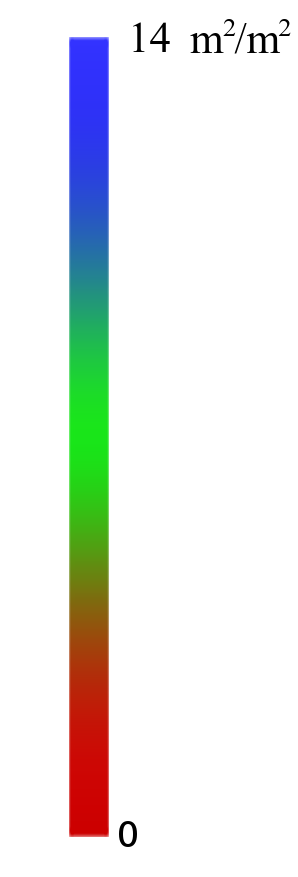}
\end{subfigure}
\caption{Waite vineyard, integrated canopy density side view captured at top: operational speed (5-6 km/h), bottom: slow speed (1-2 km/h). Total row length is 91 m. The upper row in each view has medium/sparse density and the lower row has dense canopy. Coloured red - green - blue, with maximum integrated canopy density at 10.4 $m^2/m^2$.} 
\label{fig:fastslow1}
\end{figure*}

We scan the University of Adelaide's experimental vineyard at their Waite Campus (Waite vineyard) once at the operational speed of 5-6 km/h, and once at 1-2 km/h. The slower speed is considered a more accurate scan due to the denser measurements generated with the spinning lidar. In addition, to demonstrate the robustness of our method with respect to vine density, we include one sparse and one dense vineyard row. Figure~\ref{fig:fastslow1} shows the side view of two rows (one sparse, one dense) at two different speeds (top: operational, bottom: slow). 

\begin{figure}[t]
\centering
\includegraphics[width=.5\linewidth]{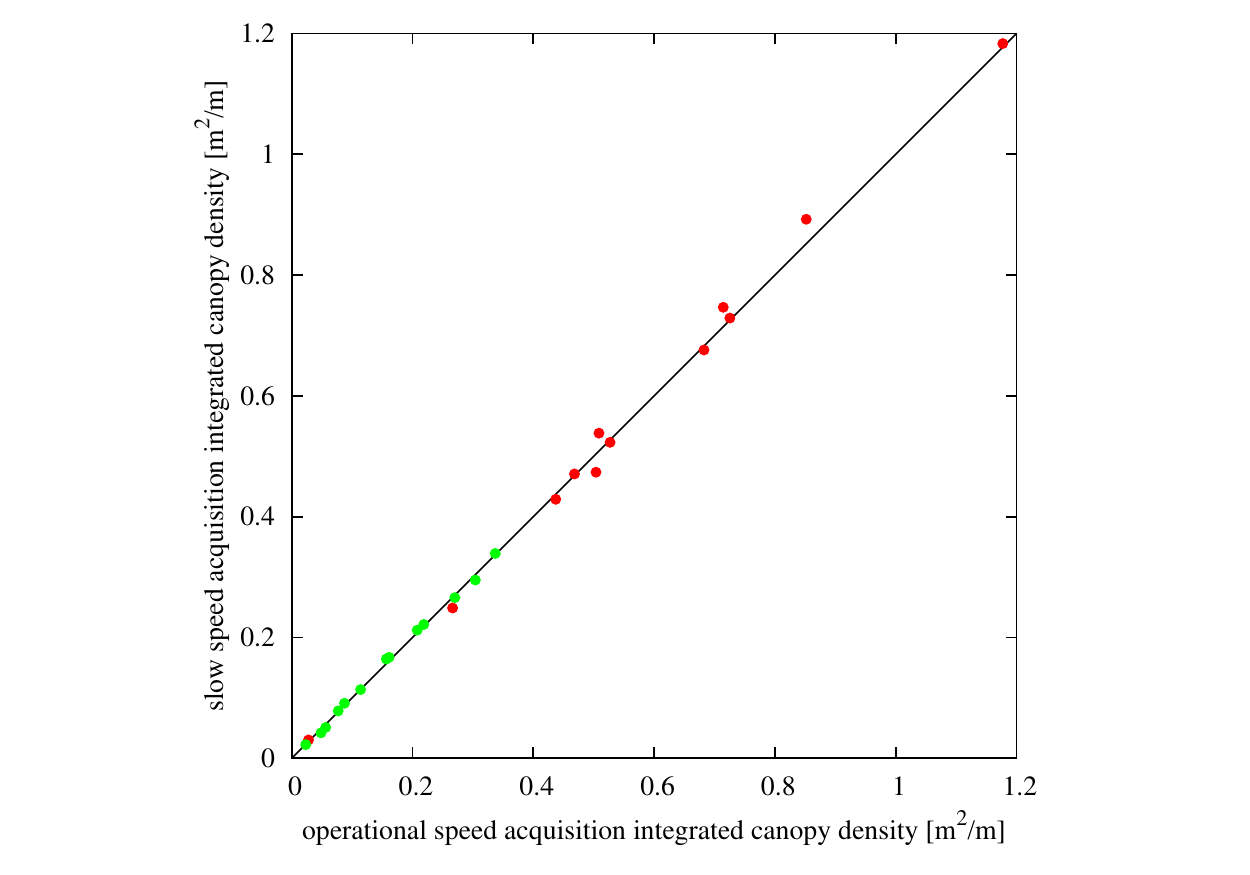}
\caption{Waite vineyard, integrated canopy density (leaf area per metre) for each 7 m long panel, comparison of operational speed (5-6 km/h horizontal axis) to slow speed (1-2 km/h vertical axis), totalling 26 panels. RRMSE is 3.4\%. Results for two rows, one of which is medium/sparse (0 to 2.5 $m^2/m$ green) and one dense (0 to 10 $m^2/m$ red), corresponding to the top and bottom row of Figure~\ref{fig:fastslow1} respectively. }
\label{fig:fastslow}
\end{figure}

To provide quantitative comparison, we measure the integrated canopy density (leaf area per metre along the row) for each panel. Each panel is approximately 7m long, resulting in a total of 26 panels. We repeat this process for each row (spare, dense), and compare these estimates at the two different speeds in Figure~\ref{fig:fastslow}.  

The canopy density estimation for the operational speed scan matched that of the lower speed scan within a Relative RMSE (RRMSE) of 3.4\%. RRMSE is the Root Mean Square Error between the two samples, divided by the mean value of all the samples. This low RRMSE demonstrates the invariance of our method to differing acquisition speeds. Also, the use of two different row densities demonstrates the estimation accuracy over range of densities.

\subsubsection{Repeatability Experiment}

\begin{figure}[t]
\centering
\includegraphics[width=0.5\linewidth]{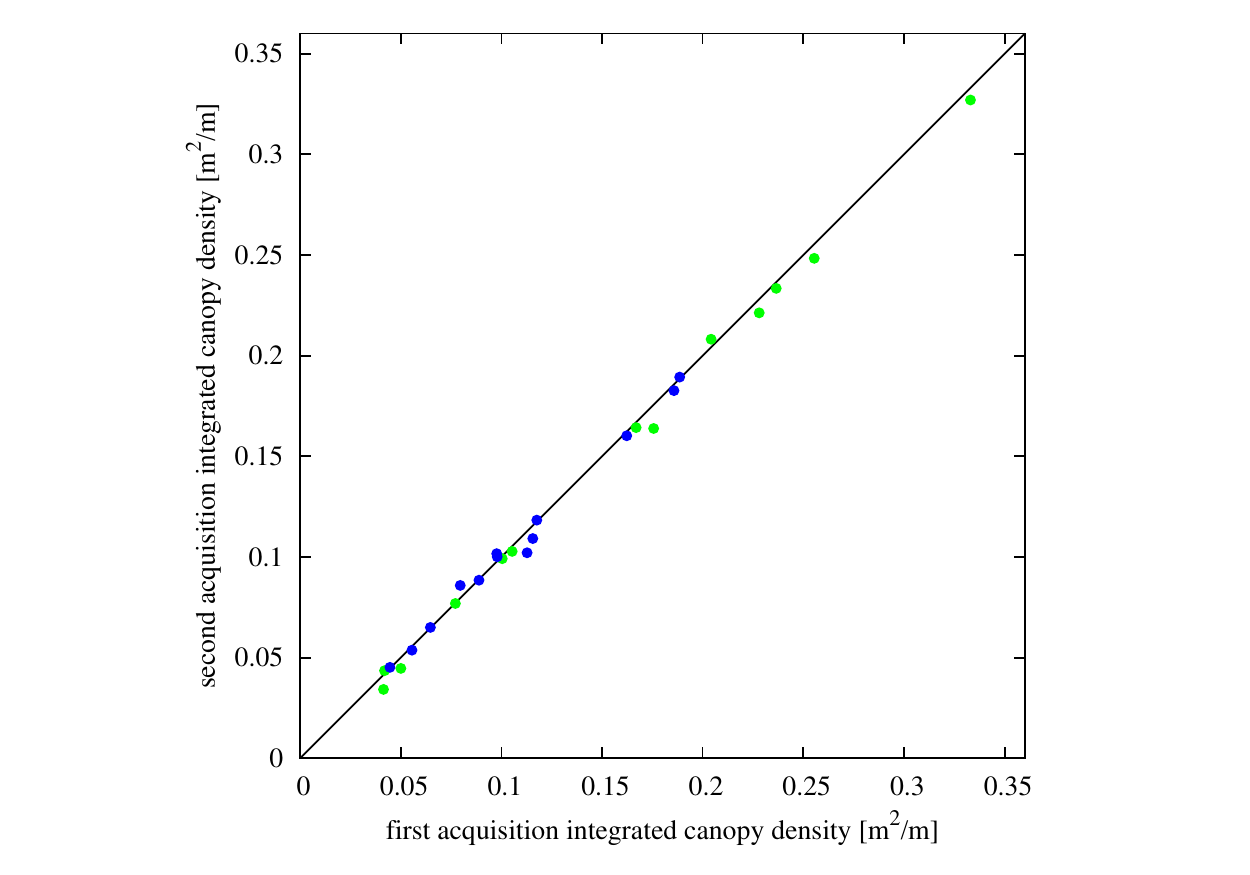}
\caption{Waite vineyard, leaf area density comparison of two successive scans (one per axis) of the same vineyard at the operational speed of 5-6 km/h. Shown for 26 panels. RRMSE is 3.8\%, the two scanned rows (blue and green) correspond to the top and bottom row of Figure~\ref{fig:fastfast1} respectively.}
\label{fig:fastfast}
\end{figure}

\begin{figure*}[t]
\centering
\begin{subfigure}[b]{0.88\textwidth}
  \centering
  \includegraphics[width=.99\linewidth]{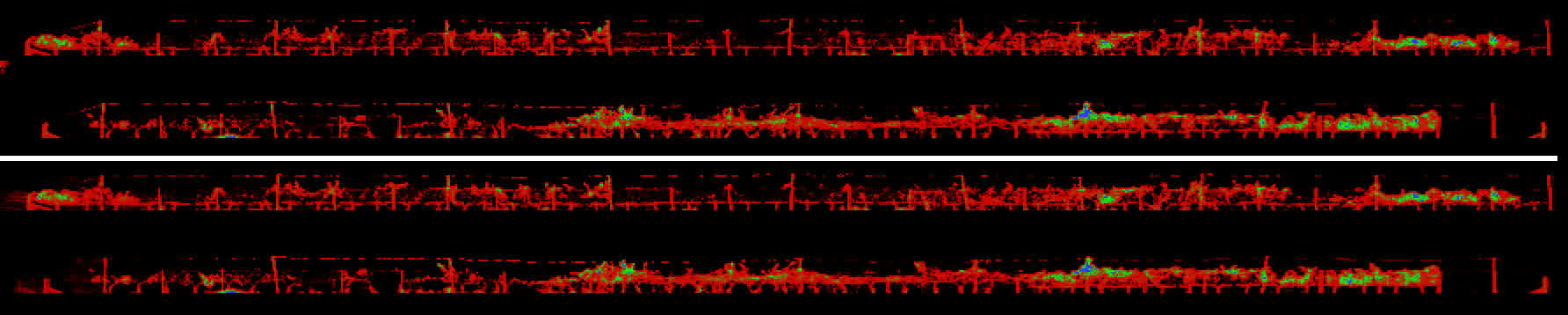}
\end{subfigure}
\begin{subfigure}[b]{0.066\textwidth}
  \centering
  \includegraphics[width=.99\linewidth]{figures/colorbarLAI.png}
\end{subfigure}
\caption{Waite vineyard, integrated canopy density side view captured at operational speed (5-6 km/h). Both rows show medium/sparse canopy density. Coloured red - green - blue, with maximum integrated canopy density at 10.4 $m^2/m^2$.} 
\label{fig:fastfast1}
\end{figure*}

To demonstrate repeatability of our solution, we compared two successive measurements of two rows of the same vineyard at the operational speed of 5-6 km/h. The estimated canopy density, per metre along row for each 7m long panel, falls within the RRMSE of 3.8\%, see Figure~\ref{fig:fastfast1}. The variation is likely due to minor deviations in the estimated trajectory and speed yet the estimated canopy densities are highly repeatable. Our qualitative evaluations are given in Figure~\ref{fig:fastfast}. It is illustrated that the two separate scans are almost identical. 

\subsubsection{Dynamic Estimation Experiment}

In this experiment we demonstrate dynamic canopy density estimation from one block of the Mclarenvale vineyard at five dates over its annual growth cycle. Firstly just after pruning in August 2018, then in November and December 2018, then just prior to harvesting in February 2019 and post harvest in March 2019. 

\begin{figure*}[]
\begin{subfigure}[b]{0.5\textwidth}
  \hspace{-10mm}  
  \includegraphics[width=1.2\textwidth]{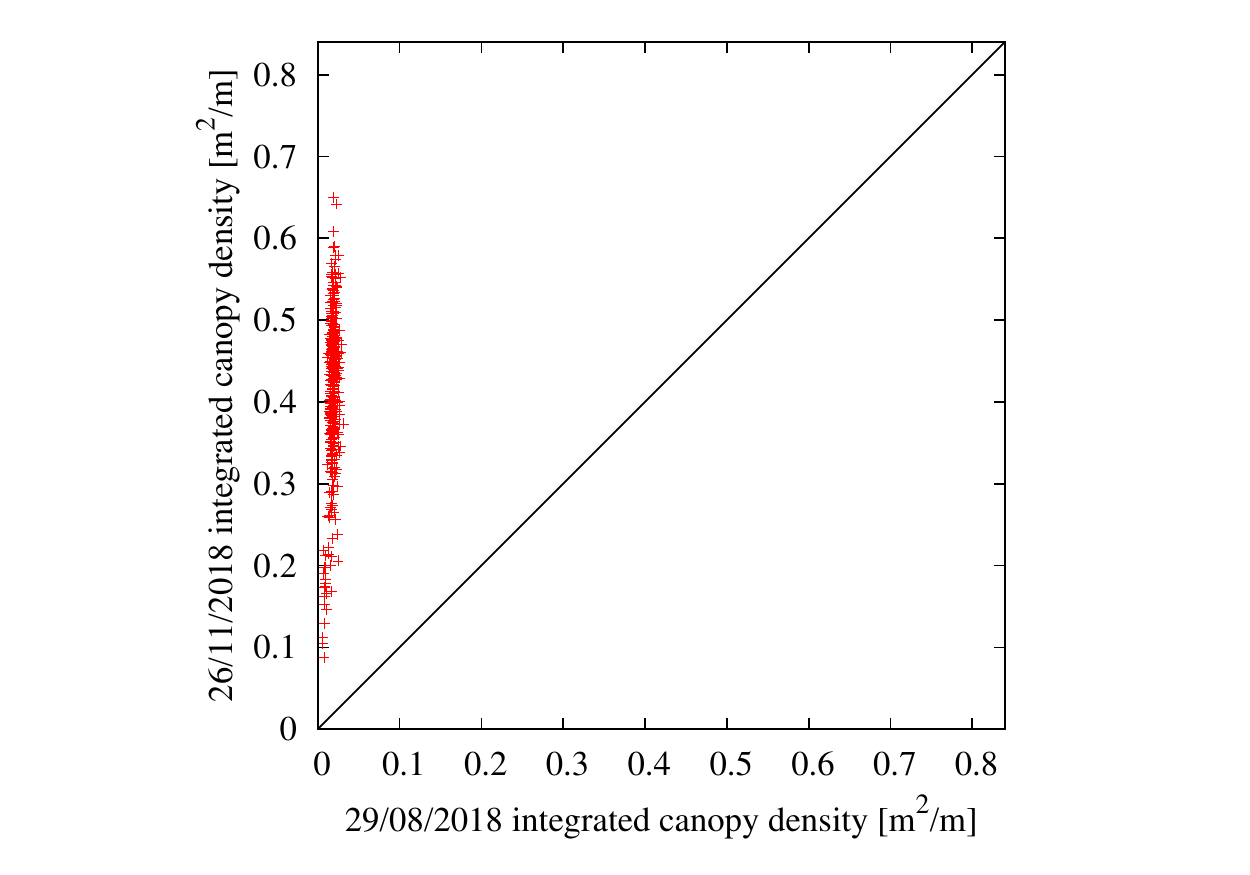}
  \caption{}
\end{subfigure}%
\begin{subfigure}[b]{0.5\textwidth}
  \hspace{-10mm}  
  \includegraphics[width=1.2\textwidth]{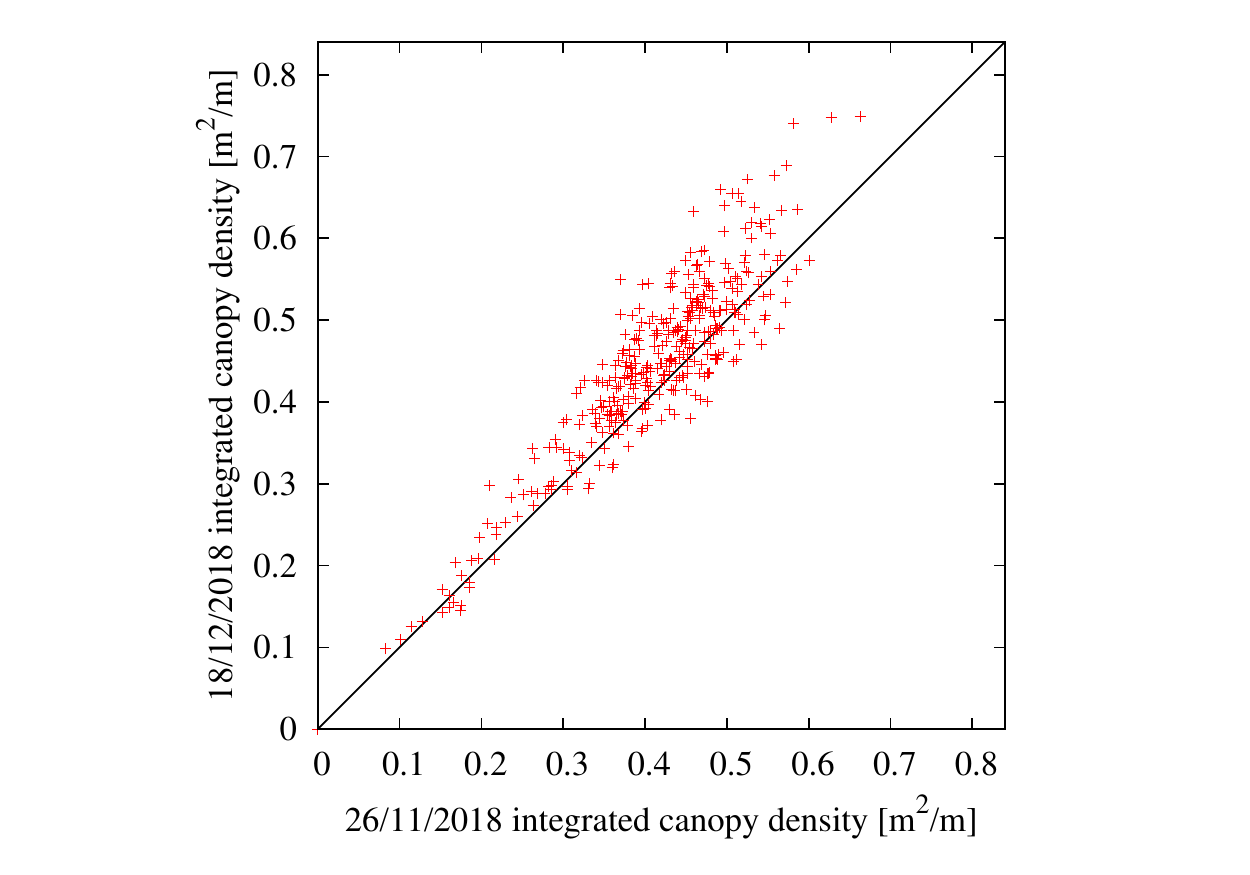}
  \caption{}
\end{subfigure}
\begin{subfigure}[b]{0.5\textwidth}
  \hspace{-10mm}  
  \includegraphics[width=1.2\textwidth]{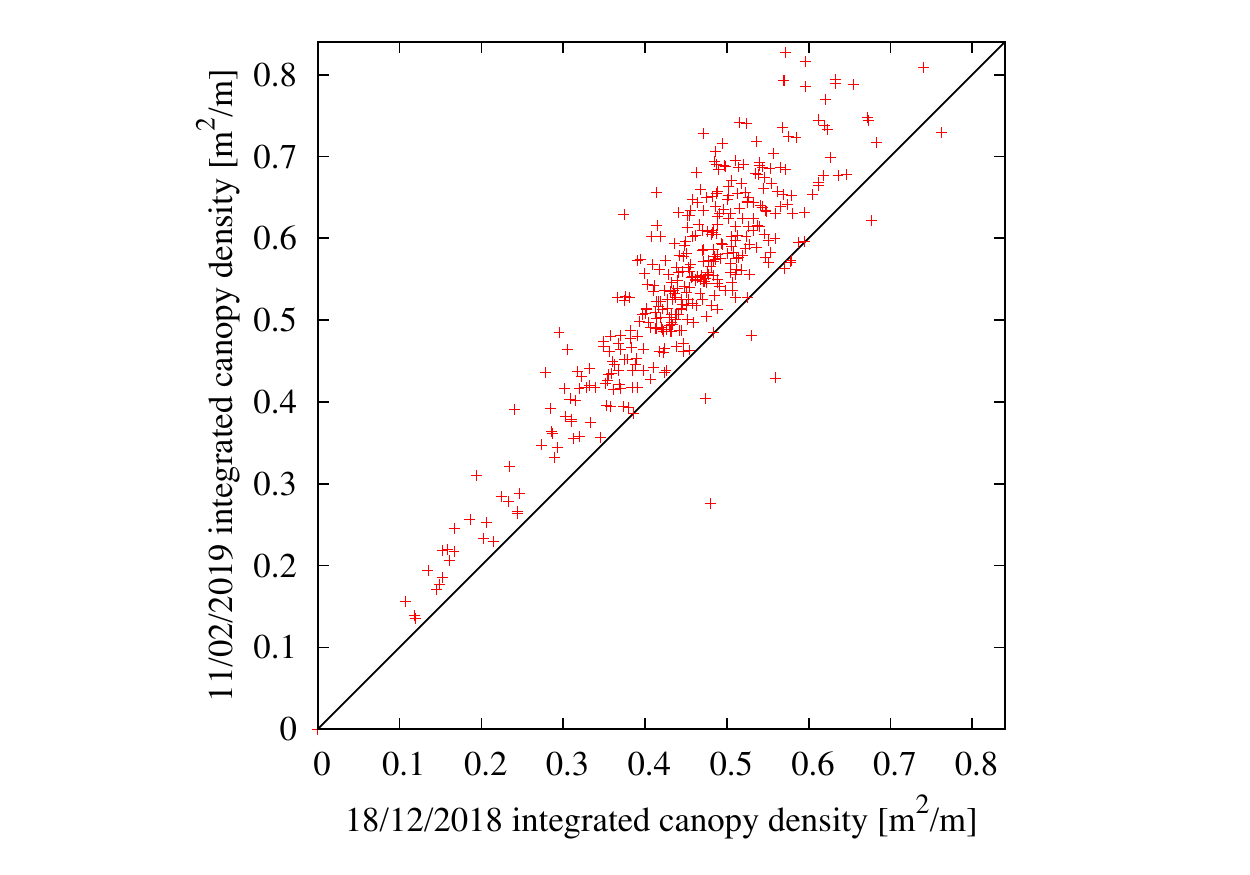}
  \caption{}
\end{subfigure}%
\begin{subfigure}[b]{0.5\textwidth}
  \hspace{-10mm}  
  \includegraphics[width=1.2\textwidth]{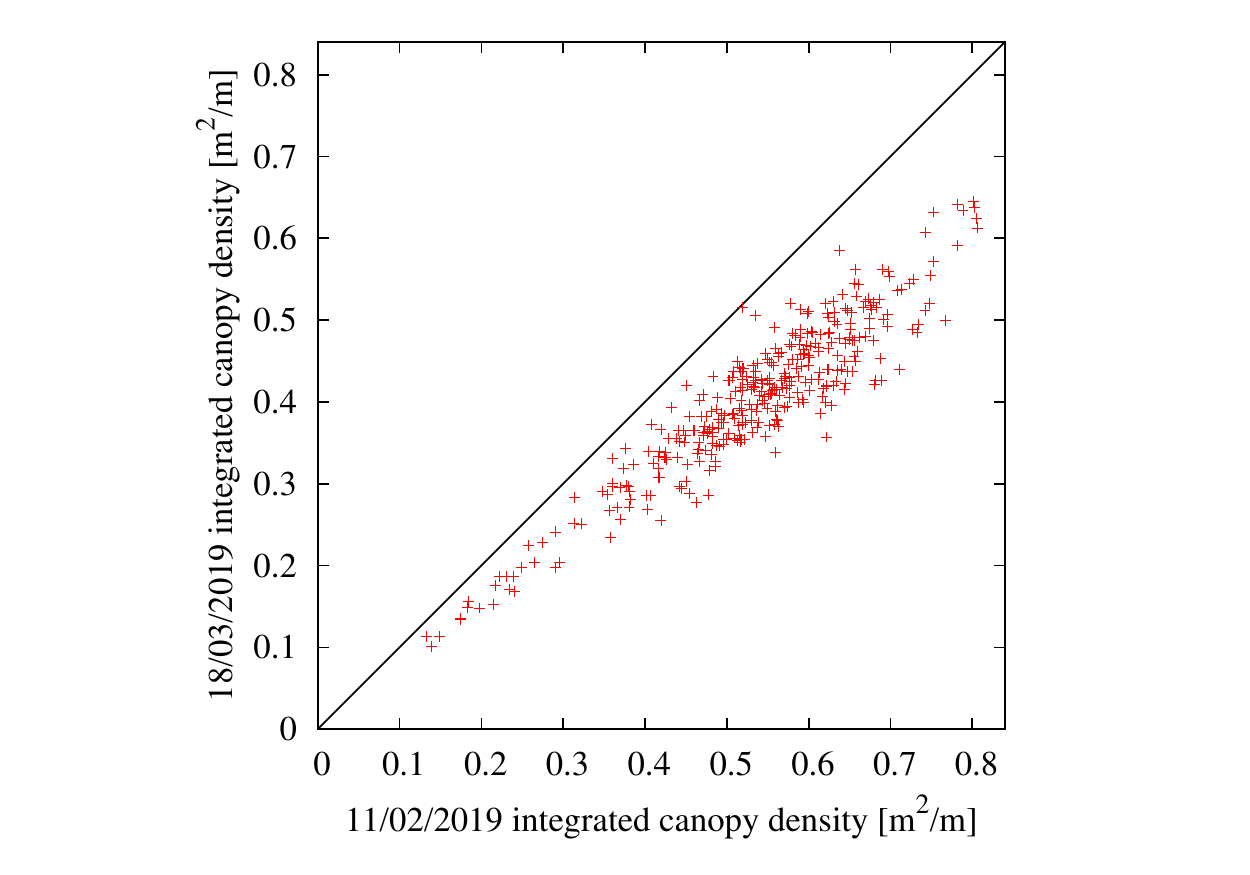}
  \caption{}
\end{subfigure}
\caption{Pairwise comparison of integrated canopy density (leaf area per metre along row) per panel for Mclarenvale vineyard at adjacent times from post pruning in August 2018 to post harvesting in March 2019. The canopy grows from its pruned state in August to thick vegetation at the end of February, where the vines are harvested using a shaker that causes leaves to fall off. Linear trend lines shown in blue.}
\label{fig:pairwise}
\end{figure*}

The integrated canopy density (leaf area per metre along each row) is combined into an average value per panel (the length of trellis between two posts), and these averages are compared for each pair of adjacent scan dates. The results are shown as four scatter plots in Figure~\ref{fig:pairwise} where points above the diagonal indicate leaf growth and points below indicate leaf loss.

The plots of scan date pairs show what was expected of the vineyard over its annual cycle, which is a growth approximately in proportion to the time interval between August and February, followed by a reduction in leaf area after harvesting, due to the loss of leaves by the vine shaking process at harvest. The largest below-diagonal outliers in Figure~\ref{fig:pairwise} (c) can be traced back to vine panels that have been deliberately pruned between the December and February dates.

\subsubsection{Multi-season Experiment} 

\begin{figure}[]
\centering
\includegraphics[width=0.55\linewidth]{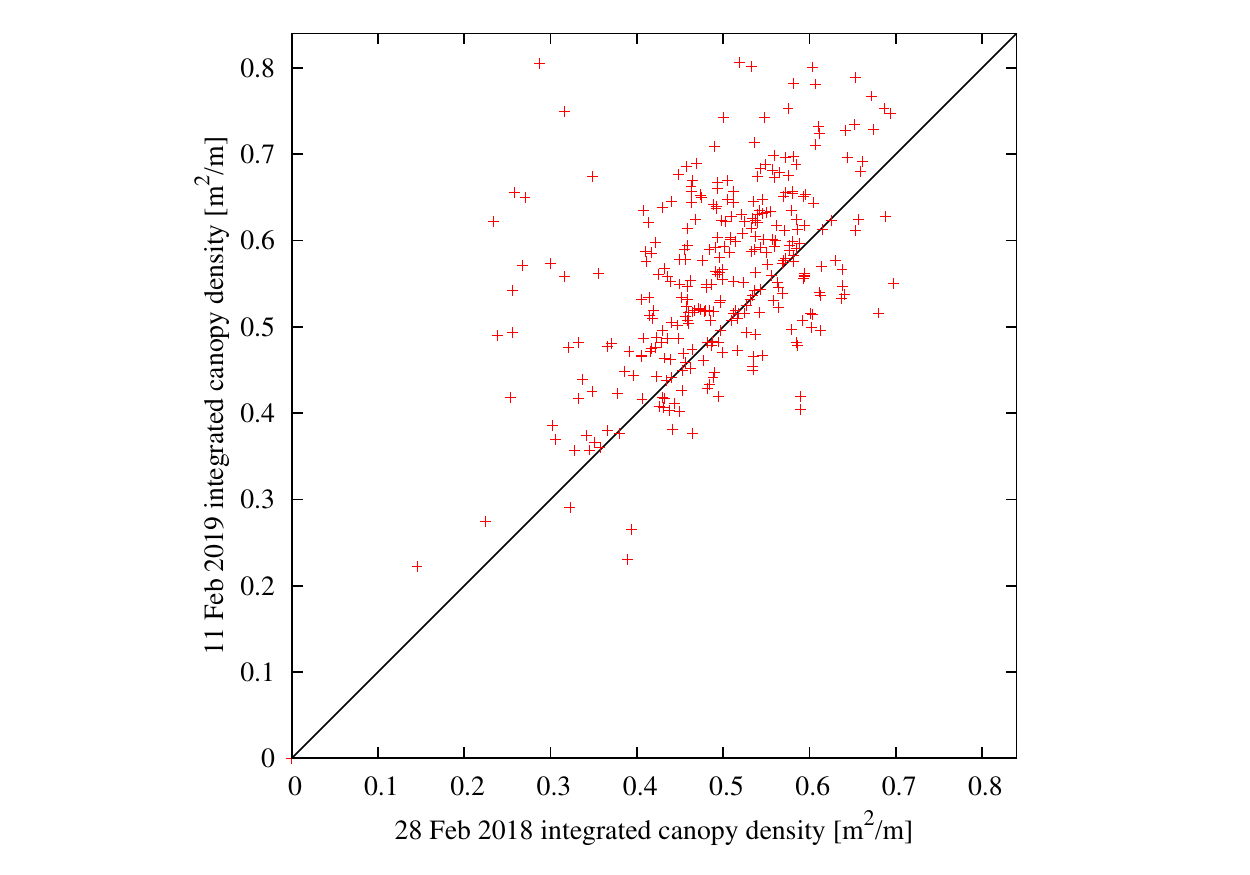}
\caption{Comparison of integrated canopy density (leaf area density per metre) at each panel, between seasons. Here February 2018 against February 2019, at Mclarenvale. This has an RRMSE of 21.5\%, demonstrating comparatively large changes in canopy structure from one season to the next. Linear trend line shown in blue.}
\label{fig:year2yearhist}
\end{figure}

To demonstrate our proposed solution across multiple seasons we run an experiment on the same vineyard \cblue{over two growing seasons.} \cblue{The estimated canopy density for February (pick season) 2018 is compared to the density in the next season, in February 2019, see Figure~\ref{fig:year2yearhist}. In this figure, one can see a larger spread than the month-apart comparisons in Figure~\ref{fig:pairwise}. In addition, the majority of points are above the above the line of equality, showing that the vineyard in 2019 had a higher density of foliage than at the same time in the previous season.}

\cblue{
\subsubsection{Precision Experiment}
From field experiments thus far, we have shown that the results from our numerical simulations transfer to the field. They are repeatable and robust to acquisition velocity, and the seasonal changes are in line with what it is expected from the growth patterns. The final experiment is to validate whether the overall density values matches an established method for estimating canopy density of grapevines. %
The most practical and established method to estimate Leaf Area Index (LAI) is by measuring gap fraction (the fraction of an image that is not considered a leaf), and using a mathematical model to infer the Leaf Area Index \cite{Miller1967}. We used the LiCOR plant canopy analyser (models Li2200, Li2250c) to extract one image per panel of the Rymill vineyard, for three rows captured in November and December 2019.

In order to compare the LAI measurements from LiCOR plant canopy analyser with our method, we integrate the canopy density within each panel (to give the one-sided leaf area per panel) and divide it by the area of ground that the panel occupies. This is the panel length (5.4 m) multiplied by the row spacing (3.0 m).

The mean LAI over both dates and both rows from LiCOR is 1.93 m$^2$/m$^2$ compared to the mean LAI 1.99 m$^2$/m$^2$ from our lidar based method. Both are calculated from a total of 306 vine panels. The results suggest the overall performance of our proposed method matches the manual LiCOR-based estimation techniques. However, our method has a standard deviation of 0.22 m$^2$/m$^2$ whereas the LiCOR-based estimation has a standard deviation more than twice as large, at 0.58 m$^2$/m$^2$. The data for each of the three rows, and the combined results are shown in Table~\ref{tab:licorcomparison}.

\begin{table}[b]
\begin{center}
 \begin{tabular}{c c c} 
 {\bf Row} & {\bf LiCOR} & {\bf Our method}  \\ 
 \hline
 28 & 1.92 (0.56) & 1.95 (0.23)\\ 
 30 & 2.05 (0.63) & 2.08 (0.21) \\ 
 32 & 1.81 (0.62) & 1.94 (0.21) \\ 
 Average & 1.93 (0.58) & 1.99 (0.22)\\ 
\end{tabular}
\end{center}
\caption{LAI estimation for three rows of Rymill vineyard, acquired in Nov and Dec 2019, comparison of mean LAI LiCOR-based estimation and lidar-based method, with standard  deviations in parentheses.}
\label{tab:licorcomparison}
\end{table}

The amount of variation demonstrates the challenges involved in determining ground truth. Gap-fraction based approaches have difficulties in delineating between panels, they can be sensitive to lighting conditions and the results are weighted towards the density of the nearest leaves. However, by taking the mean estimated LAI over 306 vine panels, we do see an alignment in the overall scale of the density values.}

\section{Discussion}
\label{sec:discussion}

\begin{figure}[!h]
\centering
\includegraphics[height=0.2\textwidth]{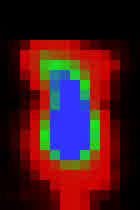}
\includegraphics[height=0.2\textwidth]{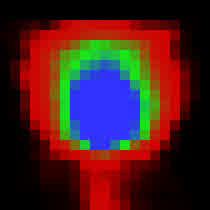}
\includegraphics[height=0.2\textwidth]{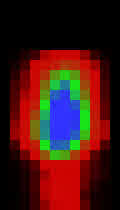} 
\includegraphics[height=0.2\textwidth]{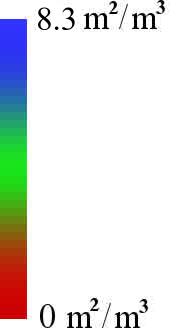}
\caption{Comparison of row shape in Feb 2019 for three vineyards, the Mclarenvale vineyard, the Rymill vineyard and the Katnook vineyard. A red-green-blue gradient is used. This is the integrated canopy density along the row for each pixel, divided by the row length.}
\label{fig:endon}
\end{figure}

\cblue{The presented results provide a weight of evidence that the per-voxel application of Eq~\eqref{eq:density} is an effective and accurate measure of vineyard canopy density. Due to the challenges in obtaining ground truth leaf area at vineyard scales, we presented a sequence of experiments to demonstrate our method's validity. Starting with simulation results, we then performed field experiments to validate the expected behaviours: repeatability, invariance to acquisition speed, applicability over a range of densities, and verifying that seasonal growth patterns are as expected. We finished by comparing results directly with an industry-standard gap-fraction based estimation, to validate the overall scale of our estimated densities.

Comparisons to gap-fraction based techniques are challenging, as the observed regions do not closely correlate to the volume of each panel, and because gap-fraction methods naturally weight density by proximity. We believe this proximity-weighting contributes to the gap-fraction based estimate having larger variance. Gap-fraction methods can also be sensitive to lighting conditions and the precise choice of sensor placement. As such, we consider the ground-truth comparison to be approximate. It nevertheless demonstrates the similarity in the mean density values. 

Now that we have covered the effectiveness of the method, we can look at several additional ways in which the data can be presented to aid the viticulturalist, beyond those already displayed.}
For situations where canopy management occurs at the block or part-block scale, the grower is likely to want to define a typical vine structure and manage the (whole) block to achieve that on average. One useful way to assess this would be to integrate estimated canopy density along the length of the row, giving an average cross-section structure per row. Fig~\ref{fig:endon} shows the visual difference between three different types of vineyard, reflecting the differing pruning and management plans. The viticulturalist could use this information to define a desired structure and use this data to assess their success and how that affected the vineyard yield and fruit quality. At present this is typically done visually and visually assessing an entire vineyard is not likely to be robust.

\begin{figure}[!h]
\centering
\includegraphics[width=0.4\linewidth]{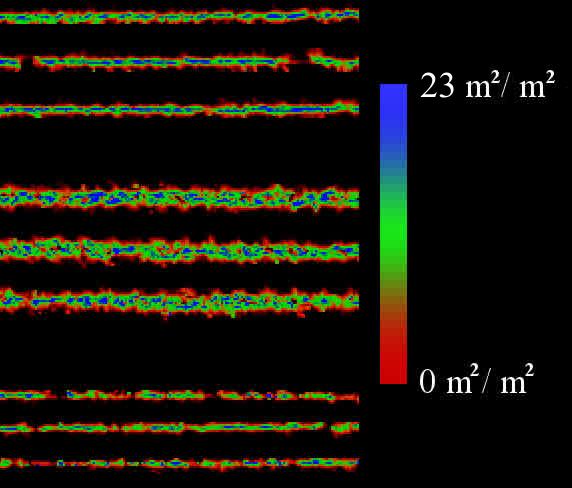}
\caption{Comparison of plan views for three rows of three different vineyards taken at the same stage, in February 2019: Mclarenvale, Rymill and Katnook. Red-green-blue gradient shows the Leaf Area Index in the vertical axis, shown in plan view. The thicknesses reflect the row shapes given in Figure~\ref{fig:endon}.}
\label{fig:topdown}
\end{figure}

The type of map traditionally used for precision viticulture is a top down view. This allows spatial variation within the vineyard to be easily assessed and the grower to make adjustments to management of sections of the vineyard, either to remove variation or to harvest different areas of the vineyard separately. Fig~\ref{fig:topdown} provides a section of such a view for the same three vineyards. Again, the difference in canopy structure is clearly evident, but now we can see the variation within the row, with gaps and failing vines easily identified. 

\begin{figure*}[!h]
\centering
\includegraphics[width=0.9\linewidth]{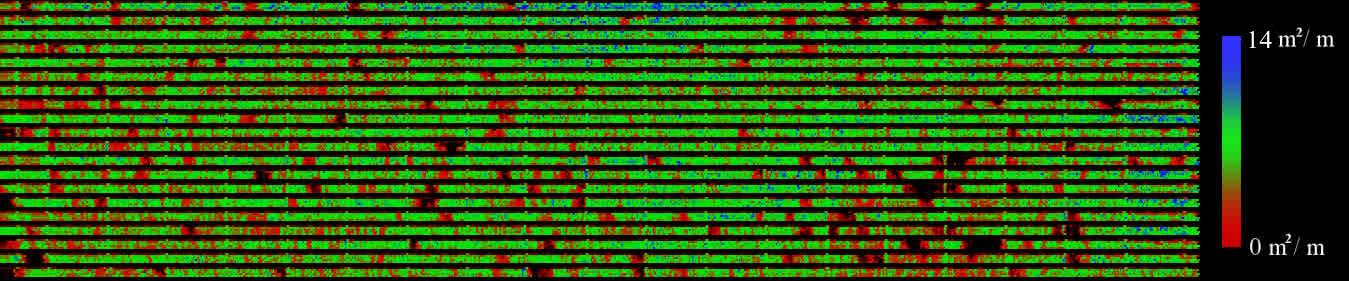}
\caption{Integrated canopy density (leaf area per metre along the row) shown in red-green-blue gradient. Each row is stacked from August 2018 at the top down to March 2019 at the row base, showing a five time-point progression. The vertical lines show the panel posts. Careful examination shows several areas of infilling, where the vines have grown between November and March, to fill several gaps.}
\label{fig:rowstack}
\end{figure*}

Visualising changes over time allows the canopy development within a season to be monitored and the management changed, depending on whether excessive or low growth is occurring. It also allows seasons to be compared, thereby determining what canopy structure was associated with what quality of fruit between vintages. Visualising this 
is more challenging, possibly the most thorough summary is to take the one dimensional canopy density along each row and compare these. For example, stacking them vertically per acquisition date over the year. Such a layout is presented in Fig~\ref{fig:rowstack}. Each vineyard row in the figure consists of four rows of pixels, one row of pixels per time point, representing a vertical history of the row's canopy density along its length. 

\section{Conclusion}
\label{sec:conclusion}
\cblue{We presented a novel canopy density estimation for perennial horticultural crops at the field scale using AgScan3D, a mobile vehicle-mounted 3D spinning lidar system, to generate a globally registered ray cloud.} Once the globally registered ray cloud is generated, a series of automated extraction and segmentation algorithms are applied to remove ground surface and segment individual rows. The accuracy of the canopy density estimation was verified to within 10\% using Monte Carlo simulation of triangular leaf distributions. Field experiments were conducted at four vineyard sites in South Australia, which varied in vineyard structure and vine management. Data were collected over two growing seasons (23 months), \cblue{64 data collection campaigns} resulting in a total traversal of 160 kilometres and 42.4 scanned hectares of vines with a combined total of approximately 93000 of scanned vines. We demonstrated  generalization, robustness, repeatability and precision of our proposed method in a series of specific experiments.

A number of different reductions of the data were then presented, to demonstrate how to convert the large amount of data into a more useful and focused array of measurements, this is particularly necessary when analysing data over a large number of time periods, as we have presented. Our experiments show canopy density repeatability of 3.8\% (Relative RMSE) per vineyard panel, for acquisition speeds of 5-6 km/h, and under half the standard deviation in estimated densities when compared to an established industry standard gap-fraction based solution.

In addition to the spinning lidar being required for accurate localisation and mapping of the vineyard, we also demonstrated in Table~\ref{tab:one} an improvement in accuracy due to using a spinning lidar, as it makes the approximation of a spherical distribution of leaf normals relative to the ray closer when the leaves have a non-spherical distribution in the world frame. Therefore the spinning lidar is a convenient tool for localisation, mapping and canopy density estimation at the field scale. 

In order to promote the use of the ray cloud as the core data format for perennial horticulture crops analysis, we will publicly release an open source ray cloud tools library. This will include the ground surface extraction and canopy density estimation methods that were described in this paper, and will allow researchers to apply these techniques to their own data. Datasets used in this paper also will be made available.

\appendices
\section*{APPENDIX} 
\label{sec:appendix}

\section{Ground Extraction}
\label{sec:ground}

Vineyards are rarely grown on flat terrain. The optimal combination of altitude, aspect and soil type often preferences hilly areas. But in order to be able to compare rows quantitatively, it is important that they are all represented on the same flat plane. In order to subtract the variable ground height we need to extract the ground terrain as a surface. As we are using a spinning lidar, the ground terrain is captured in the globally registered ray cloud, and so we need to extract this ground surface without underfitting and overfitting to the data.     
Previous work in perennial horticultural crops have estimated the ground using the vehicle location~\cite{underwood2016mapping,llorens2011ultrasonic}, the lowest branching point~\cite{chakraborty2019evaluation}, and by learning to identify the ground as a mixture of Gaussians~\cite{cheein2015real,milella2015self}. Unlike these indirect methods, we directly extract the ground surface as a mesh, so provide a globally consistent surface from which to normalise all measurements. 

\begin{figure*}[t]
\centering
\begin{subfigure}[b]{.5\textwidth}
  \centering
  \includegraphics[width=.9\linewidth]{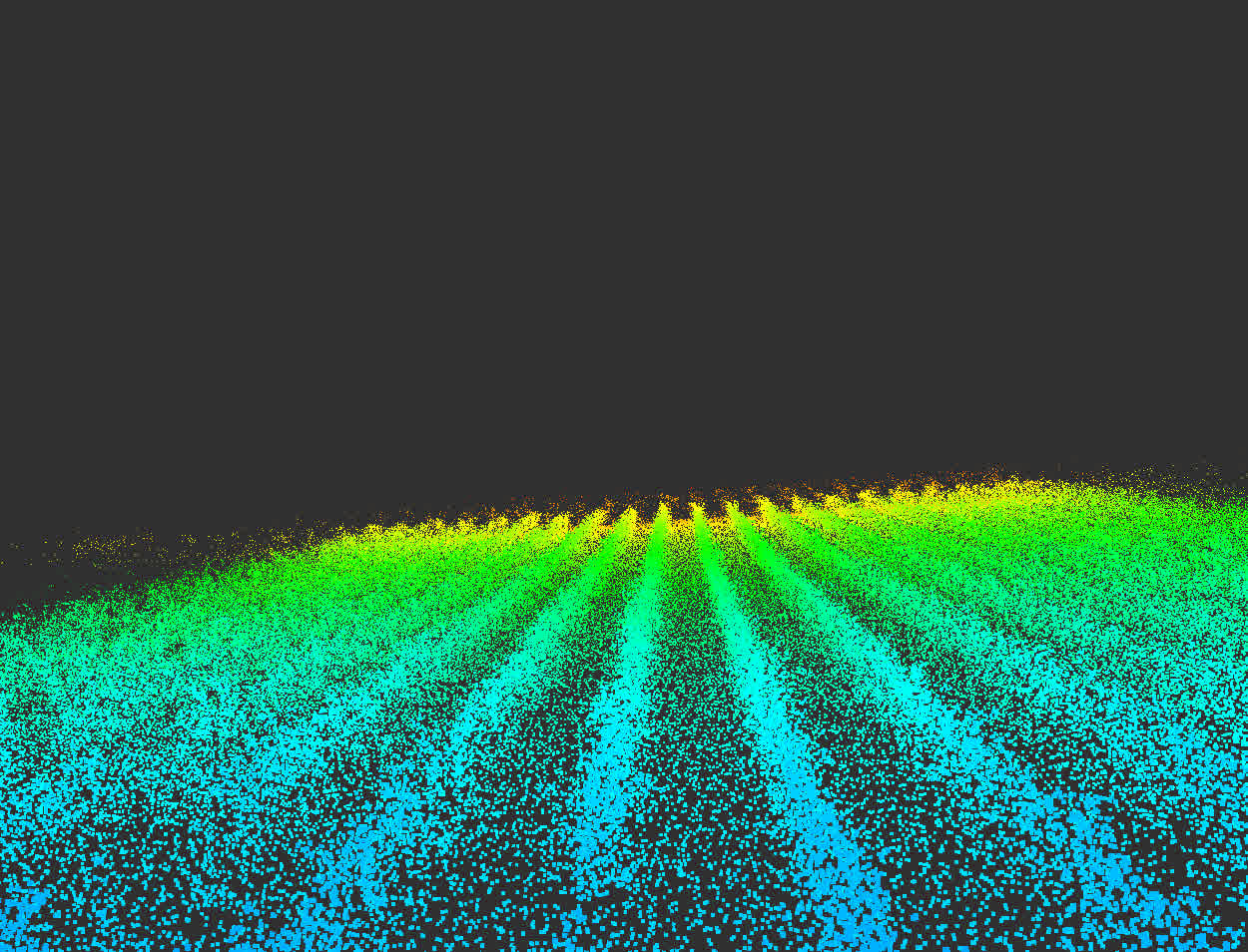}
  \caption{Ray cloud of sloping vineyard}
  \label{fig:n3_run}
\end{subfigure}%
\begin{subfigure}[b]{.5\textwidth}
  \centering
  \includegraphics[width=.9\linewidth]{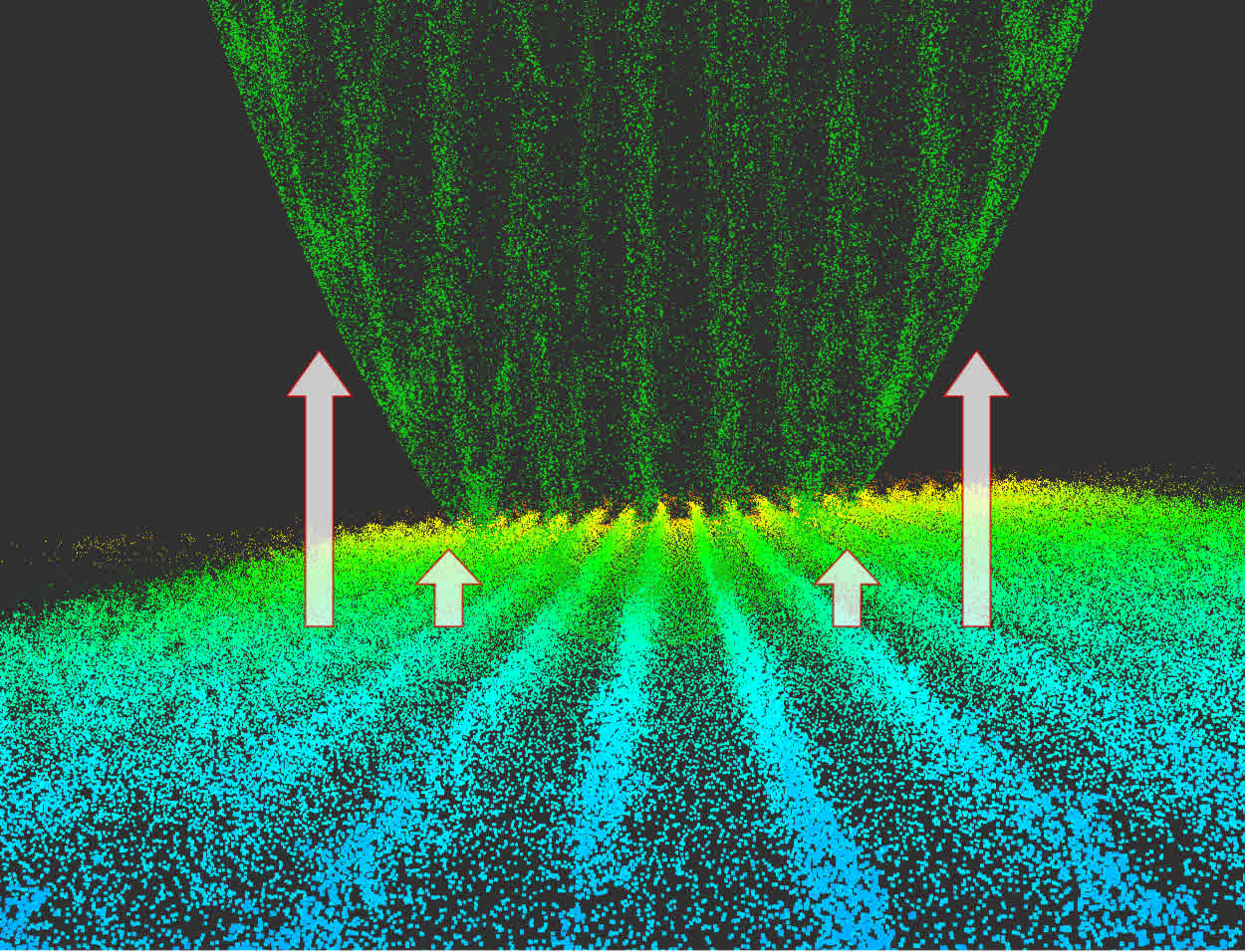}
  \caption{Add paraboloid to point heights}
  \label{fig:n3_run}
\end{subfigure}
\begin{subfigure}[b]{.5\textwidth}
  \centering
  \includegraphics[width=.9\linewidth]{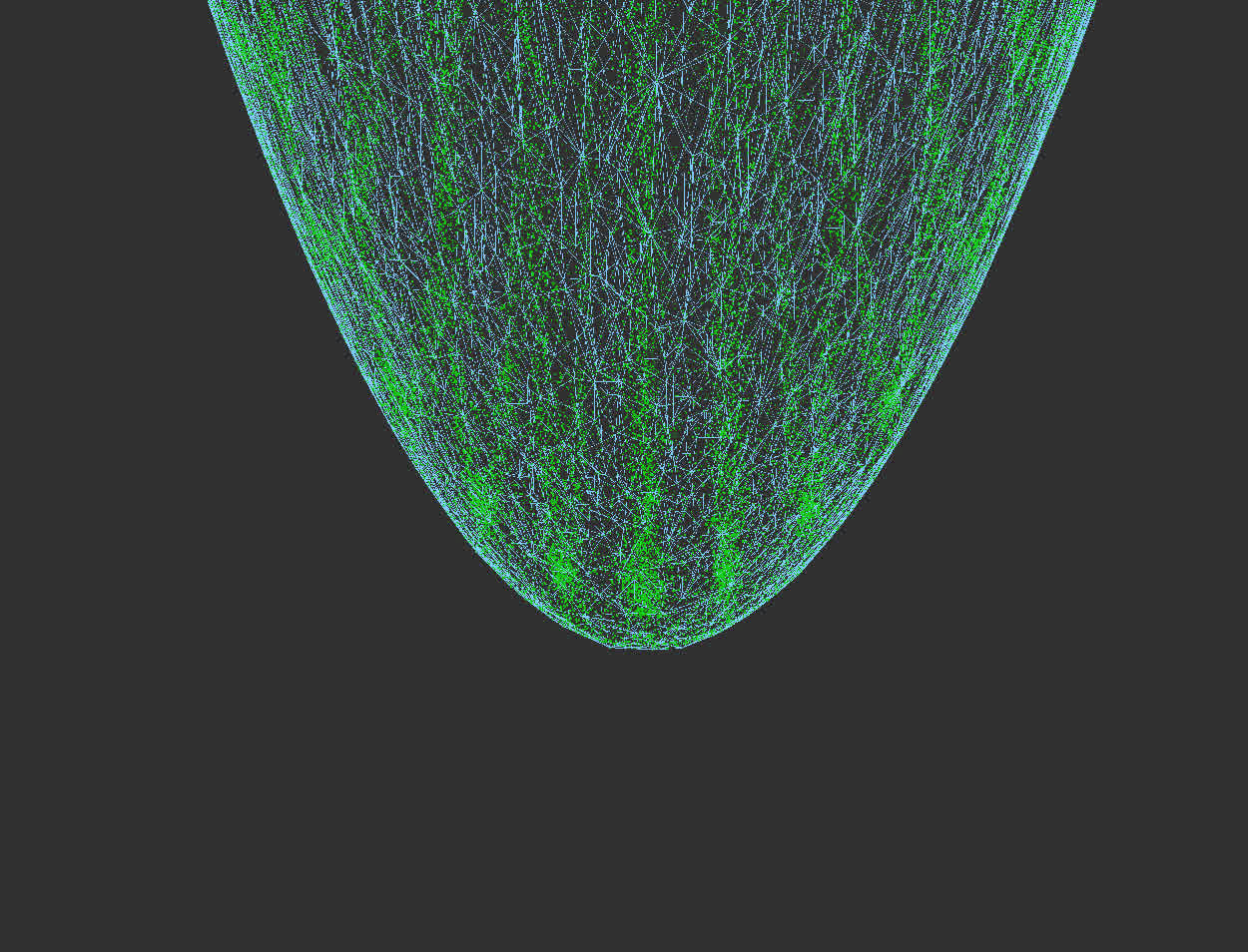}
  \caption{Extract lower convex hull (blue)}
  \label{fig:n3_run}
\end{subfigure}%
\begin{subfigure}[b]{.5\textwidth}
  \centering
  \includegraphics[width=.9\linewidth]{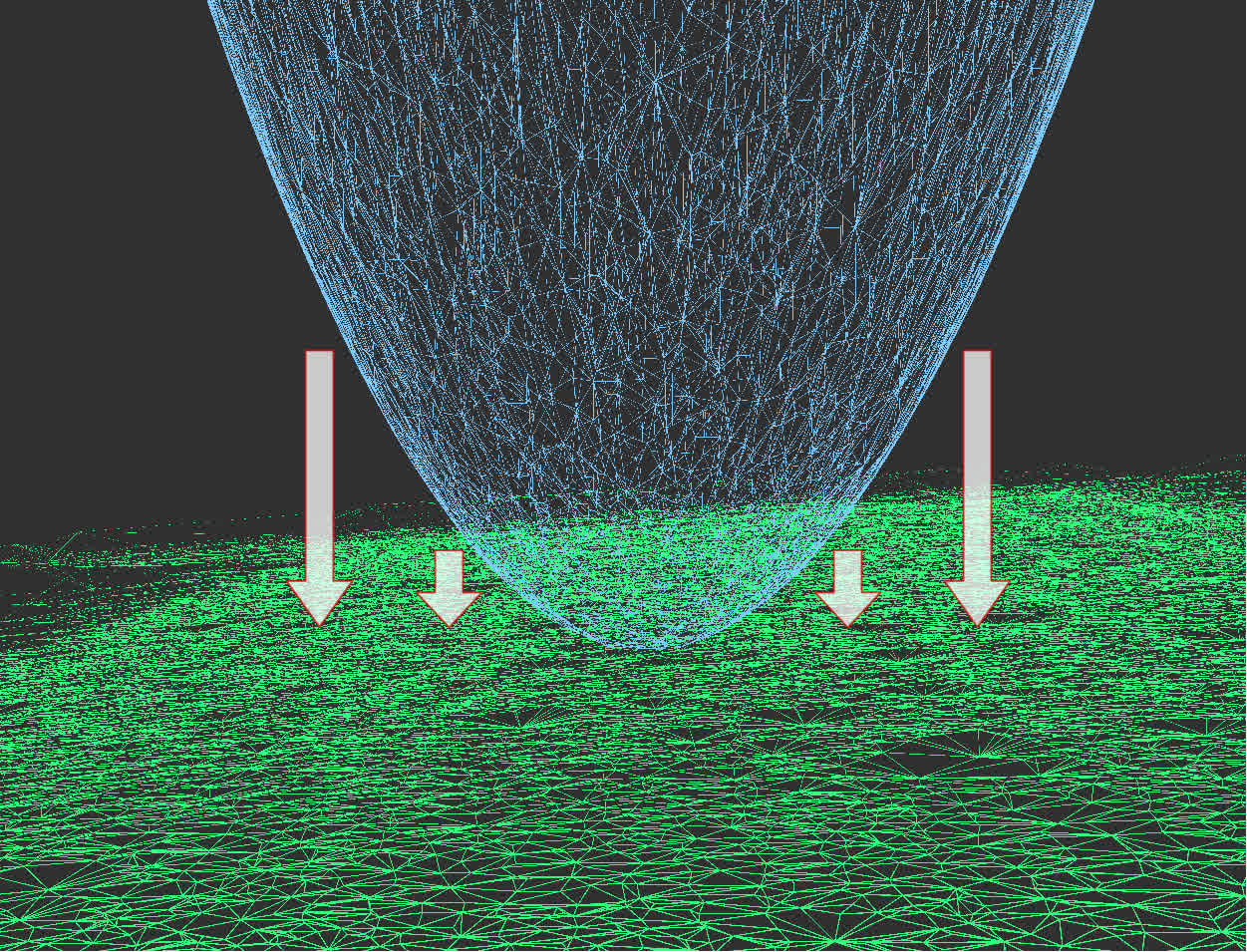}
  \caption{Subtract paraboloid from mesh vertices}
  \label{fig:n3_run}
\end{subfigure}
\caption{The proposed automated ground extraction. Four steps to extracting lower bound mesh of ground terrain, shown on a vineyard.}
\label{fig:ground_extraction}
\end{figure*}

\cblue{We propose a new ground estimation method, that is} inspired by the visibility culling method of Katz et al~\cite{katz2015visibility}. A paraboloidal function is added to the vertical component of the ray cloud end points. The convex hull of these points is computed as an indexed mesh using QHULL, and the upper surface removed. The paraboloidal function is then subtracted to give a triangle mesh that approximates a lower bound of the terrain surface, as shown in Figure~\ref{fig:ground_extraction}. 

\cblue{Let us define $X=\{\vec{p}_0, \vec{p}_1,...,\vec{p}_n\}$ to be the set of $n$ end points of a ray cloud. The function Conv($X$) converts this into a convex hull mesh, which consists of a set of $m\le n$ vertex points $\{\vec{p}_i\}$ and set of triplets defining the vertex indices of the triangles in the mesh $\{(a,b,c): 0<a,b,c\le m\}$. The ground mesh is then evaluated as $P^{-1}(\mathrm{Conv}(P(X))$ where the paraboloid function $P$ acts on each vertex or point $\vec{p}$ in its argument, according to: 
\begin{equation}
    P(\vec{p}) = \vec{p} + (0,0,k(\vec{p}_x^{\;2} + \vec{p}_y^{\;2}))^\top
\end{equation}
where $k$ is the curvature of the paraboloid base.} This single convex hull operation provides the ground mesh for the whole ray cloud, and this mesh is invariant to the horizontal location of the paraboloid.

The curvature of ground bumps that are included in the mesh is specified by the curvature of the paraboloid, we use a value of $k$=0.1 m/m$^2$\cblue{, which is sufficient curvature to fit to the sorts of undulations observed in our field trials. There is typically a large curvature range between underfitting and overfitting the surface, so we have found this single value to be sufficient for all of our scanned vineyards.} In order to subtract this ground surface height from the ray cloud we must find the triangle underneath each ray-end point and subtract the triangle face's height at that horizontal location. For this to occur efficiently we bucket the triangles into a horizontal grid of bins. Then each end point only considers the triangles in its grid cell when calculating the ground height.

\section{Row Segmentation}
Once the ground terrain is removed, we need to extract each row of the vineyard. This stage has two components, firstly we estimate the row direction as a horizontal vector, then we use this vector to segment the ray cloud into one ray cloud per each row.

\begin{figure}[H]
    \centering
    \includegraphics[width=.5\linewidth]{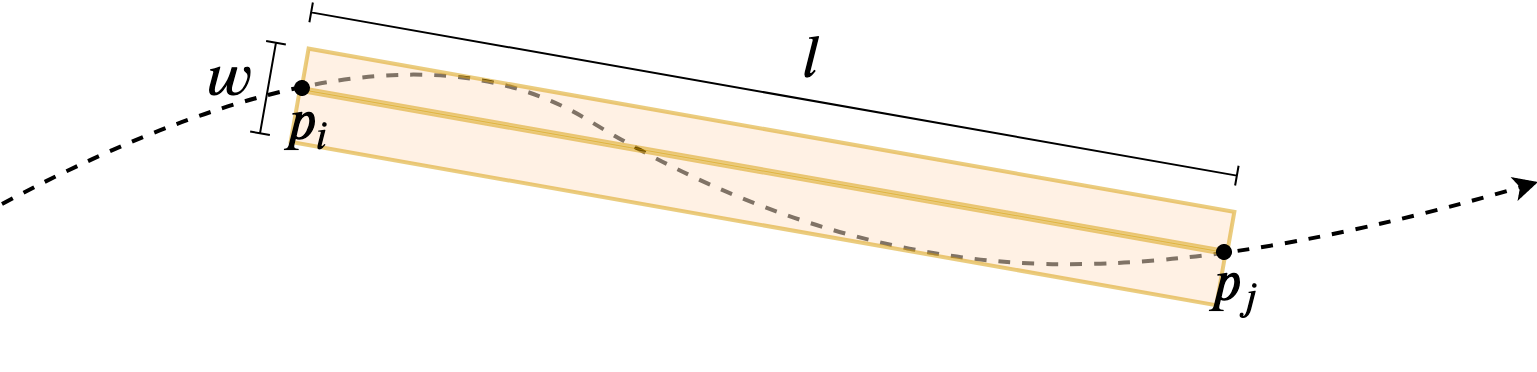}
    \caption{Showing the aligned $l\times w$ rectangle for a particular choice of indices $i,j$ along the trajectory of the sensor $p$, discretised over a short time delta. We find the $i,j$ corresponding to the largest value of $v$.} %
    \label{fig:longstraight}
\end{figure}

Row direction is found as the longest straight path of the AgScan3D trajectory while driving up and down the rows. This trajectory is the list of start locations of the ray cloud rays. 
We define straightness as the length ($l$) per width ($w$) of the bounding rectangle aligned between the head and tail points of any given segment of the trajectory. So in order to find a segment that is both long and straight, we need to maximise the product of straightness and length as $v=\frac{l^2}{w}$,
where the dimensions $l$ and $w$ are illustrated in Figure~\ref{fig:longstraight}.
We can find the approximately maximal value of $v$ over all trajectory segments using a greedy algorithm that repeatedly advances either the head or tail index along the trajectory, depending on which has the higher value of $v$.  
Once the row direction is estimated, the start locations of every ray in the ray cloud can be converted into a histogram of point densities perpendicular to this direction. The histogram is expected to show peaks corresponding to the driving location between rows. For each peak, we look at its neighbourhood down to half of this peak value, if it contains a higher peak then we throw it out, this leaves just the principle peaks. 

These principle peak locations represent the split points, from which we divide the full ray cloud into one ray cloud per vineyard row. Long rays may belong to more than one of these ray clouds. 
At this stage we can convert the ray cloud into local `row coordinates', with $x$ axis along the width, $y$ axis along the length and $z$ axis giving height relative to the ground. To localise the rays to the ground surface mesh, an accurate subtraction would transform each ray into a piece-wise linear curve, which computationally costly. So instead we approximate the localisation by subtracting the correct height from the ray-end location and the same amount from the ray-start location. This is a reasonable approximation as the most accuracy is usually required closest to the end point. 

\section{Voxelisation}
For each row we create a cuboidal grid of voxels, from 30cm in height up to the 97th percentile of end points, and in width from the 2nd to 98th percentile. The exact dimensions of the cuboid are adjusted slightly to contain an integer number of cubic voxels of width 12 cm in each dimension.

We now compute the list of rays that intersect each voxel. This can be performed efficiently by using a 3D line drawing method to iterate through the intersecting voxels of each ray in turn. For each voxel we record the number of rays that enter the voxel $n$ and the number of rays that contact within the voxel $m$. We also store the list of lengths $y_1,\dots,y_n$ of the ray extended from its voxel entry to voxel exit point, and the list of ray penetration depths $x_1,\dots,x_n$ which are from ray-start to ray-end, and clipped to within the voxel.

Applying Eq~\ref{eq:density} to each voxel gives the full 3D grid of canopy density estimates for the row. However, the accuracy of these raw estimates can be poor deep within the canopy wherever there are low ray counts ($n$). We avoid this inaccuracy by including the rays of neighouring voxels in the density calculation of an undersampled voxel, until it includes a sufficient number of rays (until $n>=10$ in our experiments).

\subsubsection*{Acknowledgments}
This research was supported by funding from Wine Australia, the Department of Agriculture's Rural R\&D for Profit program and CSIRO. Wine Australia invests in and manages research, development and extension on behalf of Australia’s grape growers and winemakers and the Australian Government. The authors gratefully acknowledge the help of staff in the Robotics and Autonomous Systems Group of CSIRO Data61 and the Winegrape and Horticulture group of CSIRO Agriculture and Food. Special thanks goes to Ross Dungavell, David Haddon, Stephen Brosnan, Emili Hernandez, Najid Pereira-Ishak for their support in this project on Hardware design, user-interface, data collection and implementations. Finally, the authors thank the vineyard owners and managers at Accolade Wine, Rymill Wine, Katnook Estate and SARDI Nuriootpa for their assistance and access to their properties.

\bibliographystyle{apalike}
\bibliography{bibliography.bib}

\end{document}